\pdfoutput=1

\documentclass[11pt]{article}

\usepackage{EACL2023}
\usepackage{graphicx}
\usepackage{dsfont}
\usepackage{times}
\usepackage{latexsym}
\usepackage{booktabs}
\usepackage{amsmath}
\usepackage{amsthm}

\usepackage[T1]{fontenc}

\usepackage[utf8]{inputenc}

\usepackage{microtype}

\usepackage{inconsolata}

\usepackage{amsmath,amsfonts,bm}

\def\eqref#1{equation~\ref{#1}}

\def\1{\bm{1}}

\DeclareMathAlphabet{\mathsfit}{\encodingdefault}{\sfdefault}{m}{sl}
\SetMathAlphabet{\mathsfit}{bold}{\encodingdefault}{\sfdefault}{bx}{n}

\def\fF{\mathcal{F}}

\def\pP{\mathcal{P}}
\def\bB{\mathcal{B}}
\def\dD{\mathcal{D}}
\def\lL{\mathcal{L}}
\def\nN{\mathcal{N}}
\def\iI{\mathcal{I}}

\def\concortname{\textsc{Repina}}
\def\concort{\textsc{Repina}$_{\text{I}}$ }
\def\capcort{\textsc{Repina}$_{\text{MLP}}$ }
\def\cnrt{RP$_{\text{I}}$ }
\def\cprt{RP$_{\text{M}}$ }
\def\stdpp{STD++ }

\def\firdivmeasure{\text{GM}}
\def\secdivmeasure{\text{HM}}
\def\zprej{z_{pre}^j}
\def\zfinj{z_{fin}^j}

\def\rR{\mathbb{R}}
\def\nN{\mathbb{N}}

\newtheorem{lemma}{Lemma}
\newtheorem{theorem}{Theorem}

\title{\emph{Representation Projection Invariance} \\  Mitigates Representation Collapse}
\author{Anastasia Razdaibiedina$^{\diamondsuit}$\enspace Ashish Khetan$^{\clubsuit}$\enspace Zohar Karnin$^{\clubsuit}$ \AND Daniel Khashabi$^{\spadesuit}$\enspace Vishaal Kapoor$^{\clubsuit}$\enspace Vivek Madan$^{\clubsuit}$ \\  \vspace{-0.6em} \\ 
$^{\diamondsuit}$University of Toronto \enspace
$^{\spadesuit}$Johns Hopkins University
\enspace $^{\clubsuit}$Amazon AI \\
\texttt{anastasia.razdaibiedina@mail.utoronto.ca} \enspace \enspace
\texttt{danielk@jhu.edu} \\
\texttt{\{vivmadan, zkarnin, khetan, vishaalk\}@amazon.com}
}

\begin{document}
\maketitle
\begin{abstract}
Fine-tuning contextualized representations learned by pre-trained language models remains a prevalent practice in NLP. However, fine-tuning can lead to \emph{representation degradation} (also known as \emph{representation collapse}), which may result in instability, sub-optimal performance, and weak generalization. 

 In this paper, we propose \emph{Representation Projection Invariance} (\concortname), a novel regularization method to maintain information content of representation and  reduce  representation collapse during fine-tuning by discouraging undesirable changes in the representations. 
 We study the empirical behavior of the proposed regularization
 in comparison to 5 comparable baselines  across 13 language understanding tasks (GLUE benchmark and six additional datasets). When evaluating in-domain performance, \concortname{} consistently outperforms other baselines on most tasks (10 out of 13). We also demonstrate its effectiveness in few-shot settings and robustness to label perturbation. As a by-product, we extend previous studies of representation collapse and propose several metrics to quantify it. Our empirical findings show that our approach is significantly more effective at mitigating representation collapse.\footnote{Our code is available at \href{https://github.com/arazd/REPINA}{https://github.com/ arazd/REPINA}. }
\end{abstract}

\section{Introduction}
\begin{figure}[t]
    \centering
    \includegraphics[scale=0.31]{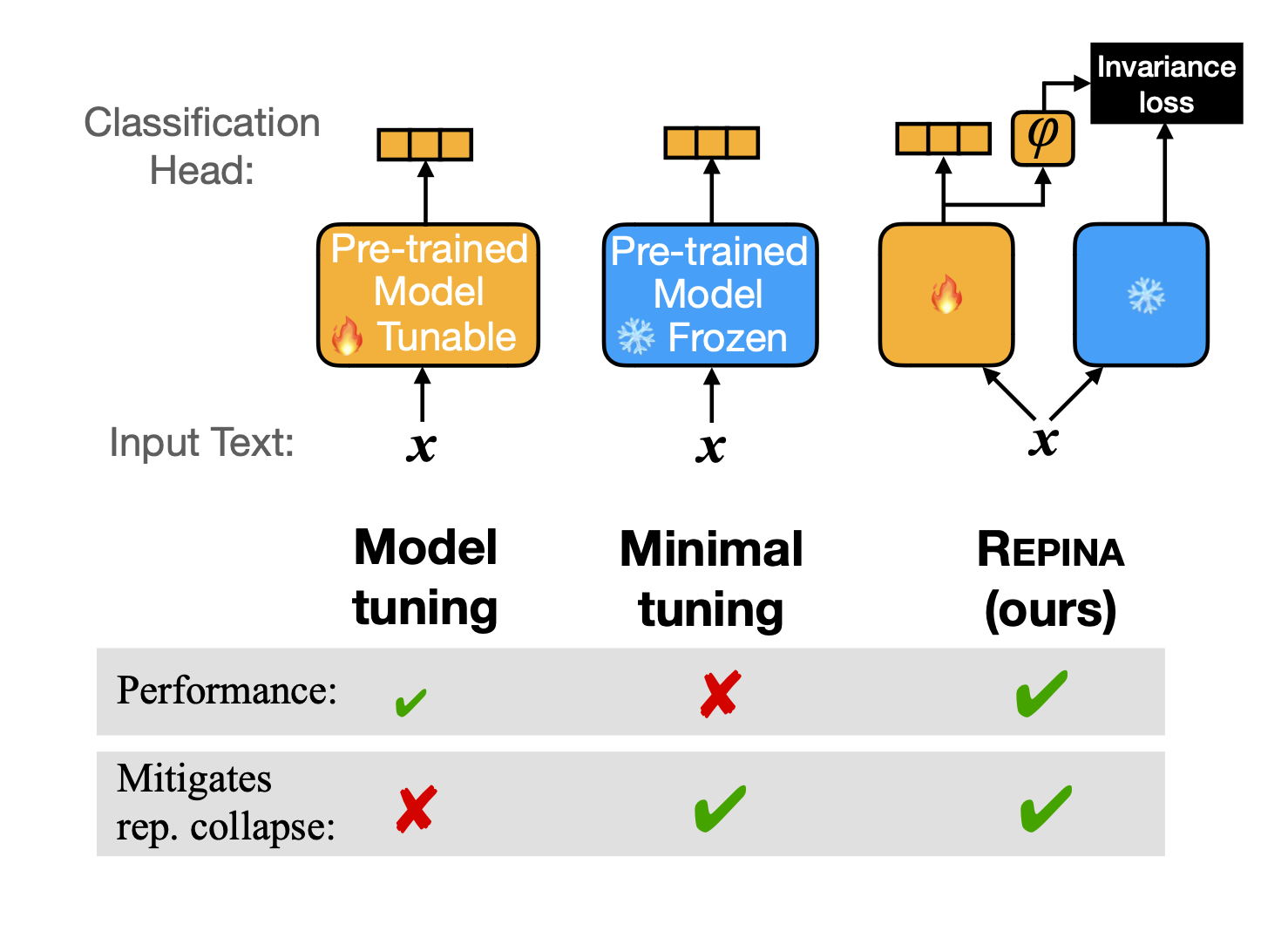}
    \caption{
      Fine-tuning whole architecture (left) generally leads to good performance though distorts the representations. 
      Minimal tuning (of classification head, for example; middle) mitigates the representation collapse but limits the model performance. 
      Our proposal \concortname{} (right) leads to good performance while mitigating the representation collapse. 
    }
    \label{fig:motivation}
\end{figure}

Fine-tuning pre-trained language models has been shown to achieve remarkable performance on a variety of natural language processing (NLP) tasks
~\cite{DevlinCLT18,brown2020language,zhang2022opt}. A standard fine-tuning strategy involves adapting the pre-trained model to a supervised downstream task (Fig~\ref{fig:motivation}; left). Such procedure can result in \textit{representation collapse}~\cite{AghajanyanZG20, zhou2021closer}, a distortion of the pre-trained representations that limits their generalizability to other domains, styles or tasks.  
An alternative approach to full model tuning is to fine-tune only several top layers, while keeping the rest of the model frozen (e.g., we could train solely a classification head, Fig~\ref{fig:motivation}; middle). 
This practice of freezing all/most of model parameters can prevent unwanted changes to pre-trained representations, but it can also limit fine-tuning and negatively affect performance~\citep{lee2019would, kumar2022fine}. This study aims to determine if it is possible to fine-tune the entire model without compromising representation quality.

\begin{figure}[tb]
    \centering
    \includegraphics[scale=0.27, trim=0cm 0.8cm 0cm 0.5cm]{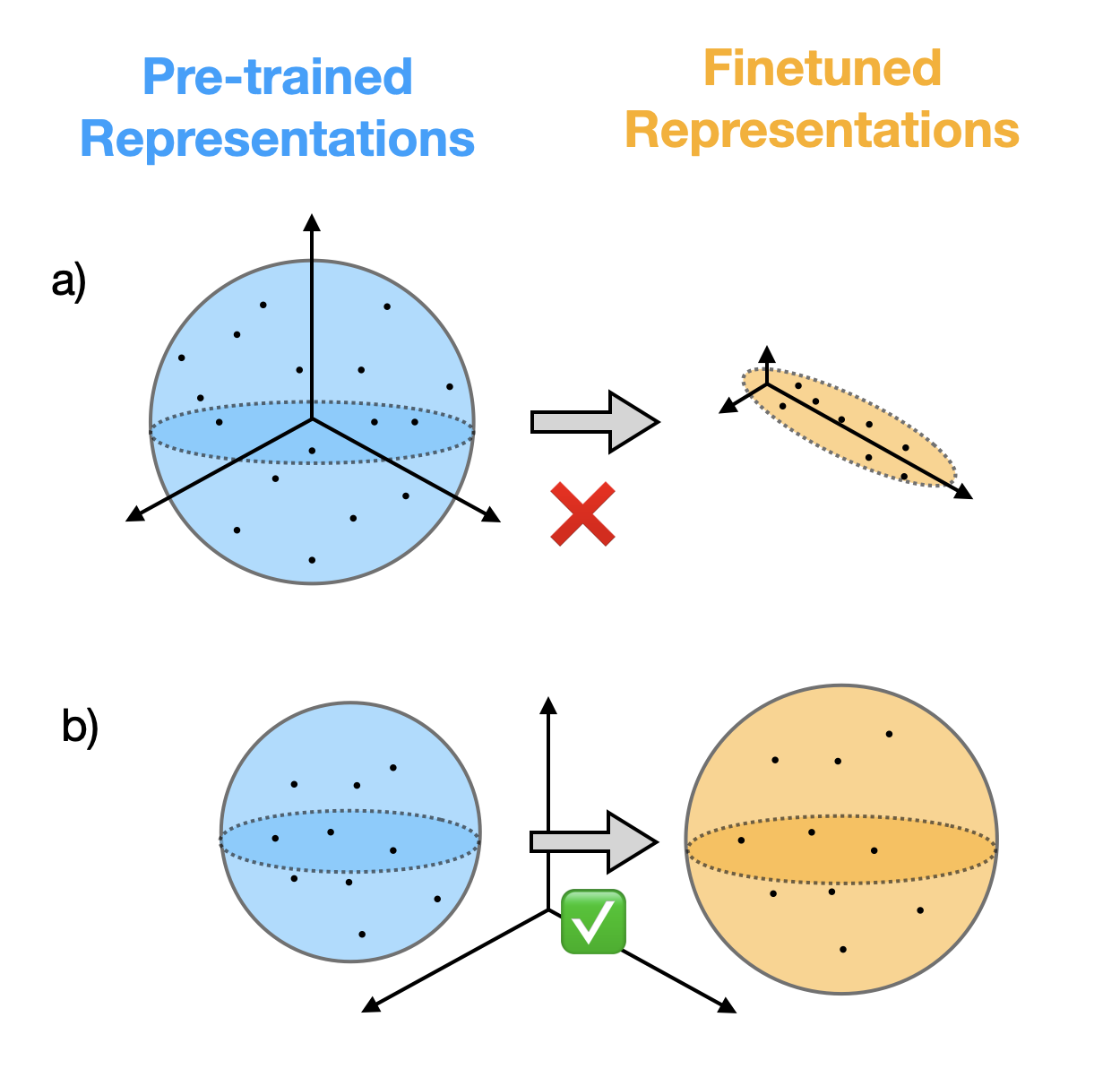}
    \caption{
    Example \textbf{\emph{a)}} shows representations \emph{collapsing} into a single dimension and losing useful information after fine-tuning. 
    Example \textbf{\emph{b)}} shows changes in representations that preserve their expressive power (e.g., coordinate shift, rotation, scaling etc.). 
    }
    \label{fig:representation_collapse}
\end{figure}

We introduce \emph{Representation Projection Invariance} (\concortname), a regularization objective that prevents undesirable changes in the representations (Fig~\ref{fig:representation_collapse}a). Our regularization applies an \emph{invariance} loss on a tunable \emph{projection} of the representation. 
In effect, this regularization allows the underlying representation to change mildly (e.g., shift and scaling) while not losing its expressivity (Fig~\ref{fig:representation_collapse}b). Our regularization objective provides a knob that controls the amount of loss-free transformations allowed during fine-tuning.

\begin{figure*}[th]
    \centering
    \includegraphics[scale=0.3, trim=0cm 0.6cm 0cm 1.0cm]{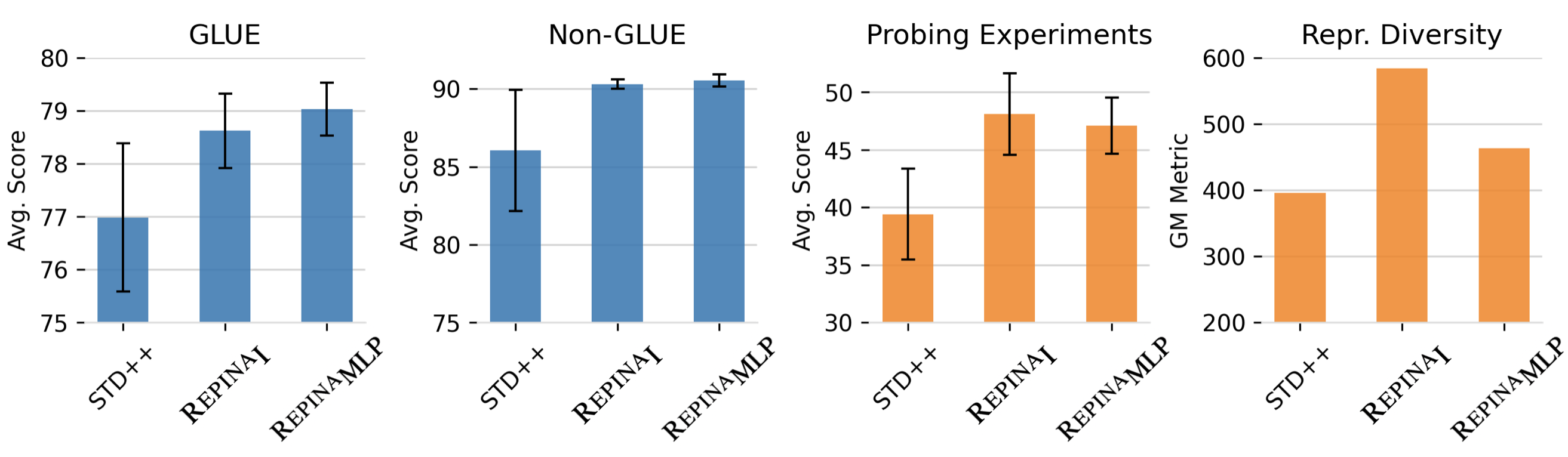}
    \caption{\concortname \space improves finetuning performance and mitigates represetation collapse. STD++: Improved variant of standard finetuning. \concort and \capcort are our methods. GLUE \& Non-GLUE: Average test set performance across seven GLUE and six non-GLUE tasks. Probing Experiments: measure of representation collapse introduced by ~\citet{AghajanyanSGGZG20} (higher is better).
    Representation Diversity: mathematical measure of the information content of representations (we report GM-5 score, see Section~\ref{sec:representation_collapse_measures}; higher is better).
    }
    \label{fig:results_summary}
\end{figure*}

We compare our method against several established regularization approaches which explicitly or implicitly address the issue of representation degradation (Section~\ref{subsection:main_results}). We show that our approach consistently outperforms major fine-tuning methods on seven GLUE classification tasks and six additional non-GLUE tasks (Fig~\ref{fig:results_summary}; left). We find that our approach is particularly effective in scenarios where data is limited (such as with only 250, 500, or 1000 examples), as the model is more likely to overfit and memorize the training data in these cases (Section~\ref{subsection:fewshot}). Furthermore, we thoroughly investigate fine-tuning under label perturbation (from $5\%$ to $30\%$ label noise) and observe that our approach is robust to incorrect labels, exceeding the performance of  standard fine-tuning procedure and common baseline methods (Section~\ref{subsection:label_noise}). 

Finally, we quantify how much different methods mitigate the degradation of representations (Section~\ref{sec:representation_collapse_measures}). We use previously explored probing experiments \citep{AghajanyanZG20}, and propose a new set of metrics that quantify representation collapse in an objective way, without requiring extra datasets/training. We observe that \concortname \space shows the strongest resistance to representation degradation among all methods.

\section{\concortname: Representation  Projection Invariance}
\label{sec:method}
Our method avoids representation collapse by preventing undesirable changes in representations during the fine-tuning process. A straightforward implementation would anchor representations during fine-tuning to their pre-trained values. That is, the final loss $\hat{\mathcal{L}}$ would combine the standard fine-tuning objective and regularizer of the deviation in representations:
{
\begin{align}
     \hat{\mathcal{L}}&=\mathcal{L} + \lambda \sum_{x \in \mathcal{I}} \|f_{pre}(x) - f_{fin}(x)\|_2^2,  \label{eq:reg_loss_init}
\end{align}
}
where $\mathcal{L}$ is the downstream task loss  (e.g. cross entropy for classification tasks), $\iI$ are the input samples of the task, $f_{pre}$ and $f_{fin}$ are the representation functions defined by the pre-trained and fine-tuned networks. Optimizing full model parameters under this  modified objective would prevent representation degradation. However, this formulation of the loss function could be very restrictive. 

There exist various transformations of a representation that maintain its expressivity (such as linear shift; Fig~\ref{fig:representation_collapse}b). 
While such transformations do not change the information content of a representation, they incur a high regularization loss based on \eqref{eq:reg_loss_init}. 

To address this issue and allow flexibility in representations while preserving their expressive capacity, we propose {\bf re}presentation {\bf p}rojection {\bf in}v{\bf a}riance regularization (\concortname): 
{
\begin{align}
    \hat{\mathcal{R}} &= \min_{\phi \in \Phi} \sum_{x \in \iI} \|f_{pre}(x) - \phi(f_{fin}(x))\|_2^2 \nonumber\\
    \hat{\mathcal{L}} &= \mathcal{L} + \lambda \hat{\mathcal{R}}. \label{eq:capcort}
\end{align}
}

Here $\Phi$ is a class of dimension preserving functions chosen before the fine-tuning process and defines the strength of regularization. 
The intuition behind the regularization objective is to incentivize the representations to be \emph{invariant} under some \emph{projection} $\phi \in \Phi$; pre-trained representations can be constructed from the fine-tuned representations by applying a function $\phi \in \Phi$.  For instance, if $\Phi$ is the set of linear functions $\{ \phi \mid \exists W,b: \;  \phi(z) = Wz + b\}$ , then we bias fine-tuned representations to be linear transformations of the fine-tuned representations.

Thus, regularization loss in the case of Fig~\ref{fig:representation_collapse}a would be zero since there exists a linear mapping from fine-tuned representations to the pre-trained representations. However, regularization loss for Fig~\ref{fig:representation_collapse}a would be high as there does not exists such a linear mapping from fine-tuned  to the pre-trained representations.

\paragraph{Choice of class of functions $\Phi$:}  $\Phi$ defines the strength of the regularizer. For instance, a singleton $\Phi$ containing an identity function is the strongest regularizer which keeps fine-tuned representations close to the pre-trained representations (equivalent to \eqref{eq:reg_loss_init}). 

Conversely, for a rich $\Phi$, e.g., deep and wide neural networks, $\phi$ can be chosen to reconstruct the lost information in  representation even if there is a severe degradation. Thus, it provides a weak regularization. Choice of $\Phi$ then depends on how prone fine-tuning is to over-fitting and how strong of a regularization method is needed. For instance, few-shot setting may require the strongest regularization and larger training datasets may require milder regularization. 

In this paper, we experiment with $\Phi$ containing identity function (\concort) and shallow multi-layer perceptrons (\capcort).  

\paragraph{Choosing the right representation:}
Normally, sentence-level representations are obtained from the final encoder blocks of a transformer model. However, it may be more beneficial to use representations from the lower layers. 
In fact, \citet{ZhangWKWA20} show that re-initializing  weights of the top layers of  encoder improves fine-tuning performance, suggesting that representation consistency may not be desirable for the top layers. Thus, we consider a variant of regularization that uses representations from the intermediate layers of  encoder. 

\paragraph{Explaining representation invariance regularization as implicitly learning multiple tasks:}

Consider the overfitting shown in Fig~\ref{fig:representation_collapse}a again. It can be prevented by fine-tuning the representations for multiple tasks simultaneously instead of a single task. This is a well known method of preventing overfitting known as multi-task learning. It not only prevents overfitting of representations but can also improves generalization performance for all of the tasks.  
 
We show that \concortname's regularization objective (\eqref{eq:capcort}) is equivalent to fine-tuning on multiple hypothetical tasks. 
Due to space constraints, we defer further discussion on the connection and the formal proof of equivalence in Appendix~\ref{sec:pseudo_meta_learning}.

\section{Related Work}
\paragraph{Mitigating Representation Collapse:} ~\citet{AghajanyanSGGZG20} study representation collapse and propose two methods, {\em R3F} and {\em R4F}, to address it. {\em R3F} induces bias towards a solution with a locally smooth prediction function, and  {\em R4F} extends {\em R3F} by adding a Lipschitzness constraint on the top classification layer. Some of the other methods which implicitly target representation collapse are {\em FreeLB} and {\em SMART}. {\em FreeLB} uses adversarial training to improve fine-tuning~\citep{ZhuCGSGL19} and {\em SMART} is a trust region-based method that avoids aggressive updates during fine-tuning~\citep{JiangHCLGZ20}.
{\em R3F}~\citep{AghajanyanSGGZG20}  has been shown to outperform both of these methods. Thus, we only include {\em R3F} in our set of baselines.  

A method that specifically targets representations during fine-tuning is {\em Supervised Contrastive Learning} (SCL).  SCL induces representations of examples with the same label to be close to each other and far from the examples of other classes ~\citep{GunelDCS20}. A major disadvantage of SCL is a requirement for large mini-batch size and, hence, heavy memory consumption. We implement a memory-efficient SCL version but exclude the original implementation from the baselines due to computational cost (see Appendix~\ref{sec:scl}). Another method which can be potentially useful for mitigating representation collapse and form part of our baseline is {\em Data augmentation}. It can be done via back translation, synonymn replacement, random deletion, and synthetic noise and is known to improve generalization performance ~\citep{FengGWCVMH21,WeiZ19}.

\paragraph{Catastrophic Forgetting}\cite{KirkpatrickPRVDRMQRG17} is a phenomenon closely related to representation collapse. In a sequential training ssetting, it refers to forgetting information learnt from previous tasks while being trained on a new task. In our context, this means forgetting the pre-training language modeling task while training for the fine-tuning task. In contrast to catastrophic forgetting, we measure representation collapse as the loss in expressive power of the representations irrespective of the performance on pre-training task. A method known to alleviate catastrophic forgetting is  {\em Weight Consolidation}~\citep{ChenHCCLY20,KirkpatrickPRVDRMQRG17}. It regularizes the fine-tuning process by encouraging fine-tuned weights to be close to the pre-trained weights. In contrast to weight consolidation, \concortname\space  {\em does not put direct constraints on weight updates, but rather tries to control the structural changes in representations.} 

Apart from working with representations instead of model weights, key difference between our method and weight consolidation is that our method does not always require fine-tuned representations to be close to the pre-trained representations. Regularizing fine-tuned and pre-trained representations to be close is a special case and the most strict form of our method (\concort).  The knob ($\Phi$) in our method can be selected to choose a regularizer ranging from this extreme case to the other extreme case with no regularization. Note that this knob is not the same as the regularization constant which is used to weigh regularization loss compared to cross-entropy loss.

Due to our limited space, we discuss further details on related works in Appendix~\ref{sec:related_work_expanded}.

\section{Experimental Set Up}
\label{sec:experiment_set_up}

\subsection{Our Methods: \concort \& \capcort}
Recall our methods \concort and \capcort introduced in Section~\ref{sec:method}. For \concort, we observe that regularizing intermediate layer representations (5th, 10th, 20th from input) perform better than regularizing the top layer (near the output, before the classifier) representation. Thus, the regularization objective for \concort is:
{
\setlength{\abovedisplayskip}{6.5pt}
\setlength{\belowdisplayskip}{6pt}
\begin{equation*}
\hat{\mathcal{R}}(\Phi = \{\mathds{1}\})=  \sum_{x \in \iI}\|f_{pre}^\ell(x) - f_{fin}^\ell(x)\|_2^2,
\end{equation*}
}

where $f_{pre}^\ell, f_{fin}^\ell$ are $\ell$-th representations from the $\ell$-th layer (from input) of the model. Choice of $\ell$ is a hyper-parameter with possible values $5,10$ and $20$. Layer 5 is most effective for small training datasets and layer 20 is most effective for large training datasets (see Appendix~\ref{sec:concort_layer_by_layer_analysis}). 

Due to computational limitations, we experiment with only top layer representations for \capcort. Thus, the regularization loss for \capcort,  $\hat{\mathcal{R}}(\Phi = \text{MLP})$ is: 
{
\setlength{\abovedisplayskip}{6.5pt}
\setlength{\belowdisplayskip}{6pt}
\begin{equation*}
\min_\Theta \sum_{x \in \iI} \|f_{pre}(x) - \text{MLP}_{\Theta}(f_{fin}(x))\|_2^2, 
\end{equation*}
}
where $f_{pre}, f_{fin}$ are the representations from the top layer of the model (before the classifier) of the pre-trained and fine-tuned model and $\Theta$ are the parameters of a multi-layer perceptron (MLP). We set the depth of MLP to 2, keeping the width equal to the representation dimension. By varying the depth from 1 to 5, we observe that for smaller training datasets, lower depth performs better.  Training with large datasets is robust to the choice of depth (see Appendix~\ref{sec:capcort_mlp_depth}).

\subsection{Baselines}
We use a diverse range of baselines for our study: 
\paragraph{\stdpp} is an improved variant of the standard fine-tuning scheme that includes the use of bias correction in AdamW, following the works of ~\citep{ZhangWKWA20,MosbachAK20} which shows that bias correction is a major cause of instability in language model fine-tuning.

\paragraph{Weight Consolidation}~\cite{KirkpatrickPRVDRMQRG17,ChenHCCLY20} is an approach that encourages agreement between pre-trained $\theta^{pre}$ and fine-tuned $\theta^{fin}$ models weights via 
a regularization term $\sum_{i} \|\theta_{i}^{fin}-\theta_{i}^{pre}\|_2^2 $ added to the loss function.

\paragraph{R3F}~\citep{AghajanyanSGGZG20} is a local smoothness regularization that prevents aggressive model updates by restricting divergence of outputs upon input perturbation. For model $f(\cdot)$ and input token embeddings $x$, R3F adds a regularization term $\text{KL}_{S} \left( f(x) \middle\| f(x+z) \right)$ to the loss function, where $\text{KL}_S$ is the symmetric Kullback-Leibler divergence and noise $z$ is sampled from a normal distribution.
\begin{table*}[ht]
\centering
\setlength{\tabcolsep}{0.35em}
\fontsize{8.6}{10}\selectfont
\begin{tabular}{llccccccccccccc}
\toprule
 Method $\downarrow$ / Task $\rightarrow$ &    RTE &   MNLI &   SST2 &   MRPC &   QNLI &    QQP &   CoLA &   Yelp &  Chem &   IMDB &  AGnews &  SciTail &  SciCite \\
\midrule
        STD++ &  70.8 &  65.6 &  92.1 &  86.8 &  87.2 &  76.7 &  59.70 &  95.3 &     82.6 &  93.2 &   91.7 &    71.6 &    81.9 \\
            DA &  73.6 &  65.5 &  92.0 &  90.7 &  87.4 &  76.4 &  \textbf{63.4} &  95.6 &     82.9 &  93.2 &   91.8 &    93.7 &    82.1 \\
           WC &  72.2 &  \textbf{66.7} &  92.7 &  88.6 &  87.2 &  76.2 &  61.5 &  95.9 &     \textbf{83.9} &  93.4 &   91.9 &    94.0 &    82.2 \\
       ReInit &  70.9 &  65.1 &  92.0 &  91.0 &  87.3 &  77.2 &  61.2 &  95.4 &     82.5 &  92.7 &   91.7 &    93.4 &    82.4 \\
          R3F &  70.4 &  65.0 &  92.1 &  89.9 &  87.0 &  74.9 &  62.0 &  95.5 &     82.9 &  93.1 &   91.7 &    86.5 &    82.0 \\
          \midrule
 \concort &  71.4 &  65.7 &  92.9 &  \textbf{91.5} &  87.5 &  79.0 &  62.3 &  95.8 &     83.5 &  \textbf{94.0} &   \textbf{92.1} &    93.7 &    82.7 \\
\capcort &  \textbf{74.4} &  65.2 &  \textbf{93.2} &  91.1 &  \textbf{87.6} &  \textbf{79.3} &  62.5 &  \textbf{96.0} &   83.7 &  93.9 &   91.9 &    \textbf{94.8} &   \textbf{83.2} \\
\bottomrule
\end{tabular}

\caption{Performance for our methods (\concortname$_{\text{I/MLP}}$) and baselines on 7 GLUE and 6 non-GLUE datasets. Average gain of 2.1 over \stdpp for GLUE datasets and 4.5 over non-GLUE datasets. \concortname beat all baseline methods in 10/13 cases. For QQP, MNLI, QNI, AGNEWS, IMDB, YELP and SCITAIL, we only used 10K training datapoints.}
\label{tab:results}
\end{table*}

\paragraph{ReInit}~\citep{ZhangWKWA20} improves fine-tuning performance by re-initializing the top-$k$ layers of the encoder (closer to the output) with gaussian random samples from $\mathcal{N}(0, 0.02^2)$. Following the original study, we perform hyperparameter search for $k=2,4$ or $6$. 

\paragraph{Data Augmentation (DA)} generates augmented samples by adding noise to the training data (keeping the label intact)~\citep{DeVriesT17}. In our implementation, we add gaussian noise $\epsilon \sim \mathcal{N}(0,\delta)$ to the token embeddings where $\delta=1e-5$.

\begin{table}[ht]
    \centering
    \small
    \begin{tabular}{cc}
    \toprule
         Method & Regularization coefficient \\
    \midrule
       \concort  & 0.01, 0.05, 0.1, 0.5\\
       \capcort & 0.01, 0.05, 0.1, 0.5\\
       DA & 0.05, 0.1, 0.2, 0.4, 0.8\\
       R3F & 0.1, 0.5, 1, 5\\
       WC &0.01, 0.05, 0.1, 0.5 \\
       \bottomrule
    \end{tabular}
    \caption{Regularization coefficient for different methods.}
    \label{tab:reg_coefficients}
\end{table}

Table~\ref{tab:reg_coefficients} show the regularization coefficients used for each method.

\subsection{Datasets}
We evaluate methods on GLUE benchmark ~\citep{WangSMHLB19} and six additional non-GLUE datasets (Table~\ref{tab:datasets}). These include: biomedical relation extraction on  CHEMPROT~\citep{KringelumKBLOT16}, 
 sentiment classification on YELP~\citep{ZhangZL15} 
and IMDB~\citep{MaasDPHNP11}, 
citation intent classification on SCICITE~\citep{CohanWMF19}, language inference on SCITAIL~\citep{TusharAP18} and article topic classification on AGNEWS~\citep{XiangJY15}. 
For each task, we use their corresponding adopted performance metric. On these 13 datasets, we conduct a variety of experiments with many and few supervision instances.

To keep the cost of fine-tuning computations on extremely large datasets (such as MNLI and QQP), we limited their training sets to $10,000$ data points, and marked with a suffix ``-10K'' henceforth. 

For datasets with no available test set labels, we use their development set to report the performance.
We use a subset of original train data split (size equal to validation set) which is not used for training for hyper-parameter selection. 

\begin{table}[ht]
\footnotesize
\centering
\fontsize{8.1}{11}\selectfont
\begin{tabular}[t]{lllll}
\toprule
    Task &  Train &   Dev & $C$ &               Metric \\
\midrule
    COLA &   8551 &  1043 & 2& MCC \\
     RTE &   2490 &   277 & 2  &          Accuracy \\
    SST &  67349 &   872 &  2  &         Accuracy \\
    MNLI-10k & 10000 &  9815 &3 & MCC \\
    MRPC &   3668 &   408 & 2   &               F1 \\
    QQP-10k & 10000 & 40430 &    2   &            F1 \\
    QNLI-10k & 10000 &  5463 &     2   &     Accuracy \\
     CHEMPROT &   4169 &  2427 &      13  &     Micro F1 \\
     SCICITE &   7320 &   916 &  3 &          Macro F1 \\
     SCITAIL-10k &  10000 &  1304 &    2 &        Accuracy \\
  AGNEWS-10k & 10000 &  5000 &   4  &        Macro F1 \\
    YELP-10k & 10000 & 10000 &    2  &       Accuracy \\
    IMDB-10k &  10000 &  5000 &   2   &       Macro F1 \\
    \bottomrule
\end{tabular}
\caption{
The datasets used in this study, their size, number of classes ($C$) and the corresponding evaluation metrics. MCC denotes Matthews correlation coefficient.  
}
\label{tab:datasets}
\end{table}

\subsection{Fine-tuning Settings} 
Due to the large scale of the experiments and in order to have a meaningful comparison with various approaches, we consistently use BERT-large model for implementing both our proposed algorithm and the baselines. Existing works such as \cite{ZhangWKWA20,MosbachAK20} also use similar experimental setups. Additionally, to verify the generality of our findings to other models,  we performed limited experiments on RoBERTa-base where observe similar performance gain. 

We fine-tune all models for 5 epochs (unless otherwise specified) at a learning rate of 2e-5, and report performance with 5 different seeds. Due to resource constraints and in contrast to prior works ~\citep{DevlinCLT18, AghajanyanSGGZG20, ChenHCCLY20}, we do not search for optimal learning rate for each method-task combination. To verify the impact of this choice, we perform limited experiments selecting best learning rate, dropout and number of epochs for each method and a subset of tasks (Appendix~\ref{sec:hpo}). We observe similar gains as reported in the main paper.

For each method, we select optimal hyperparameters by performing evaluation on the unused fraction of the training set (see Appendix~\ref{sec:experiment_set_up_appendix}). 

 Since standard fine-tuning is susceptible to failed runs that substantially lowers the average resulting performance ~\citep{MosbachAK20}, we filter out these failed runs and report average performance over 5 successful runs. We define run as failed if its performance is close to or lower than the majority classifier~\citep{DodgeISFHS20}. Majority classifier is a dummy model that always predicts the label of the  majority class in the dataset. 
 We define a threshold close to the performance of the majority classifier as per metric in Table~\ref{tab:datasets}. A fine-tuning run is a failed run if the performance on unused part of the training dataset is below the threshold. See Section~\ref{appendix:failed:runs} for the exact thresholds.

 \section{Results: Generalization Performance}\label{sec:results_generalization}
 In this section, we present experimental results with the baselines introduced in the earlier section. 
\subsection{Full dataset - Generalization performance}
\label{subsection:main_results}

Table~\ref{tab:results} shows that \textbf{\concortname{} models outperform the baselines consistently across a variety of tasks}: our method outperform other ones on 10/13 tasks. Both \concort and \capcort outperform baseline methods in terms of mean performance, with improvements in the mean performance over the corrected fine-tuning strategy \stdpp by $1.7$ and $2.0$ points respectively for GLUE benchmark, and $4.3$ and $4.5$ points for non-GLUE benchmark.

\subsection{Analyses on Fine-tuning Stability}
\label{subsec:fine-tuning:stabiltiy}
Similar to the prior literature~\cite{DodgeISFHS20,MosbachAK20, WangSMHLB19}, we observe that the standard fine-tuning procedure is prone to instability and sub-optimal convergence, leading to failed runs. Recall that we formally define a fine-tuning run as a failed run if the resulting performance is close to the majority classifier. 

\begin{figure}[ht]
    \centering
    \includegraphics[scale=0.57, trim=0cm 0.1cm 0cm 0.1cm]{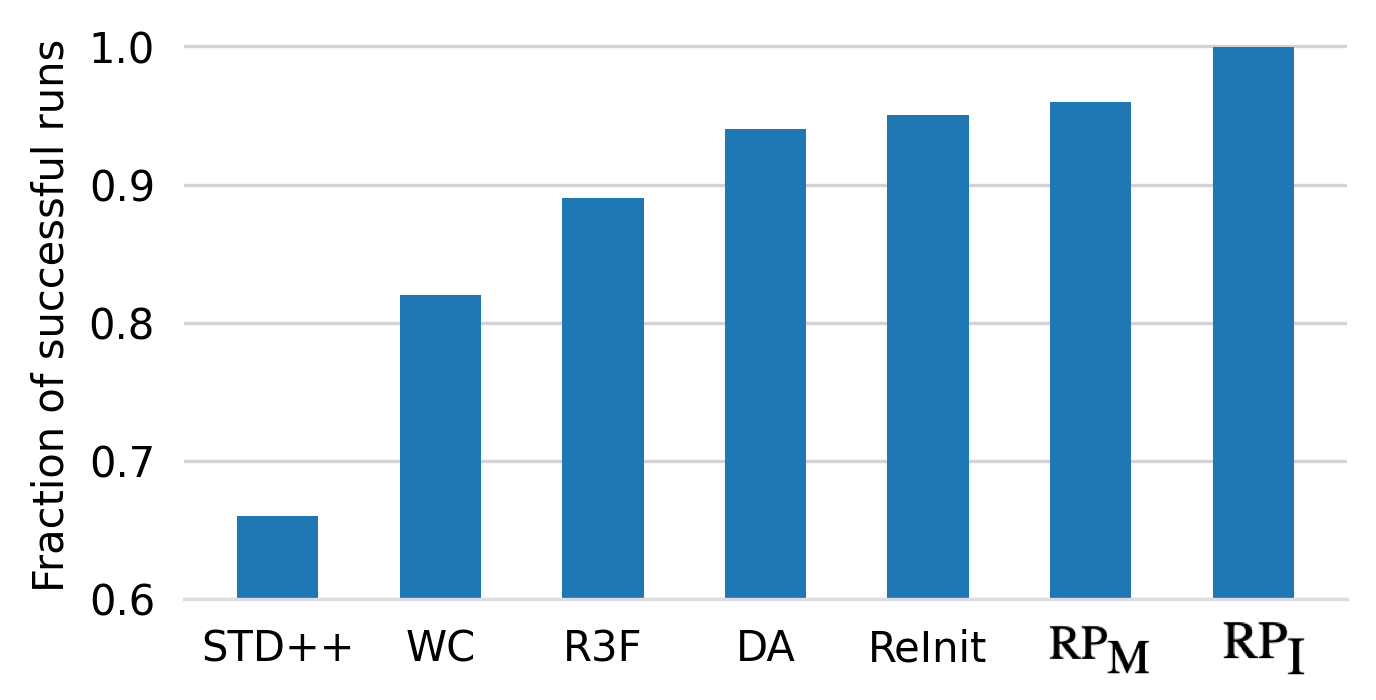}
    \caption{Fraction of successful runs across all tasks. Run is defined as successful if its test performance is higher than the performance of a majority classifier. 
    Our proposed regularization (\cnrt/\cprt) increases the fraction of successful runs, hence, leading to more stable fine-tuning behavior.  
    }
    \label{fig:successful_runs_ratio}
\end{figure}

In the previous section, we reported the mean performance of only successful runs (complement of failed runs). Figure~\ref{fig:successful_runs_ratio} shows the fraction of runs that were successful for each method.

We note that \concort has the least number of failed runs (maximum number of successful runs). Moreover, if we do not filter out failed runs, our methods perform even better than all the baseline methods. \concort achieves an average 2.6 percentage point improvement over the next best baseline method (WC). Thus, we conclude that 
\textbf{our methods demonstrate higher stability and less fraction of failed runs than other approaches}.
(additional experiments in  Table~\ref{tab:stability_results} in Appendix~\ref{sec:non_filtered_results}.)

\subsection{Comparison with parameter-efficient tuning methods}\label{sec:results_peft}

The main focus of our study is mitigating representation collapse in a full model tuning setting, which is a common way to tune LLMs since it allows to achieve the best downstream task performance, at the cost of forgetting and representation collapse. However, this goal can also be achieved with parameter-efficient tuning (PEFT) methods, which can balance good downstream performance with minimal parameter changes.

As we have discussed in the introduction, if we fine-tune only several top layers of the model, while keep-
ing the rest of the parameter frozen (e.g., train
solely a classification head, Fig~\ref{fig:motivation}; middle), representation collapse can be avoided. However, this would not lead to optimal downstream task performance. As a middle ground approach, we could train just a larger fraction of model parameters (yet not the full model) with parameter-efficient approaches. Common PEFT methods include LoRA \citep{hu2021lora}, prefix tuning \citep{li2021prefix}, prompt tuning \citep{lester2021power}, adapters \citep{houlsby2019parameter}, and variations of those \citep{ruckle2020adapterdrop, khashabi2021prompt,  lester2022reducing, razdaibiedina2023progressive, razdaibiedina2023residual}.

We provide results for prompt tuning, prefix tuning, LoRA and REPINA-MLP on 4 GLUE tasks in Table~\ref{tab:peft}. As we can see, \textbf{REPINA-MLP consistently outperforms PEFT methods}.

\begin{table}[h!]
    \centering
    \setlength{\tabcolsep}{3pt}
    \begin{tabular}{lcccc}
    \toprule
         & SST-2 & MRPC & CoLA & RTE \\
    \midrule
        Prompt Tuning & $86.4$ & $76.9$ & $59.9$ & $52.5$ \\
        Prefix Tuning & $90.9$  & $91.0$ & $57.6$ & $73.9$ \\
        LoRA & $91.5$ & $90.0$ & $60.4$ & $71.5$ \\
        REPINA-MLP & $\mathbf{93.2}$ & $\mathbf{91.1}$ & $\mathbf{62.5}$ & $\mathbf{74.4}$  \\
    \bottomrule
    \end{tabular}
    \caption{Comparison with PEFT methods.}
    \label{tab:peft}
\end{table}

\subsection{Out-of-distribution robustness}
We perform experiments on four pairs of datasets to study OOD generalization following \citet{hendrycks2020pretrained} protocol (see Table~\ref{tab:ood}). The chosen pairs of datasets induce a distribution shift, which allows to measure OOD robustness. Due to limited resources and different sizes of tasks, we limited the training set to 200 examples per class (fixed across all runs). Overall, \textbf{\concort shows steady improvement in OOD performance in all cases}.
\label{sec:ood_results}
\begin{table}[ht]
\footnotesize
\setlength{\tabcolsep}{2.5pt}
\centering
\begin{tabular}{c|cc|cc}
\toprule
Task & \multicolumn{2}{c|}{IID} & \multicolumn{2}{c}{OOD}  \\
Train $\rightarrow$ Eval & STD++ & \cnrt & STD++ & \cnrt \\
\midrule
Imdb $\rightarrow$ SST2 & $\mathbf{92.1_{\pm 0.5}}$ & $\mathbf{92.1_{\pm 0.5}}$ & $87.8_{\pm 0.9}$  & $\mathbf{88.9_{\pm 0.5}}$ \\
SST2 $\rightarrow$ Imdb & $91.1_{\pm 0.1}$ & $\mathbf{92.1_{\pm 0.1}}$ & $87.1_{\pm 0.1}$ & $\mathbf{87.8_{\pm 0.1}}$ \\ 
Yelp $\rightarrow$ Amzn & $60.7_{\pm 0.2}$ & $\mathbf{60.8_{\pm 0.9}}$ & $28.9_{\pm 0.2}$ & $\mathbf{29.3_{\pm 0.5}}$ \\  
Amzn $\rightarrow$ Yelp & $39.4_{\pm 1.9}$ & $\mathbf{41.5_{\pm 1.4}}$ & $49.7_{\pm 1.9}$ & $\mathbf{50.8_{\pm 1.4}}$ \\
\bottomrule
\end{tabular}
\caption{REPINA improves OOD performance. \cnrt: \concort; Train: training task (IID), Eval: evaluation-only task (OOD). Data is limited to 200 samples/class.  
}
\label{tab:ood}
\end{table}

\subsection{Robustness to Label Perturbation}
\label{subsection:label_noise}

Real-world data can often contain mislabeled samples, which can hinder the training process. Hence, robustness to label noise is a desirable quality of the fine-tuning approaches. Here, we study the performance of the fine-tuning methods under label perturbation. We introduce {\bf label noise} as follows: let $C = \{1,\dots,c\}$ be a class of labels and $\mathcal{X} = \{(x,y)\}$ be the true dataset for the fine-tuning task. Our fine-tuning method has access to a noisy dataset $\mathcal{X}' = \{(x,y')\}$ where $y'=y$ with probability $1-p$ and sampled uniformly from $\{1,\dots,c\}\setminus \{y\}$ with probability $p$.

\textbf{\concort and \capcort show the highest resistance to label perturbation}, retaining closest to the original performance upon introducing $5$-$10\%$ noise to labels (Table~\ref{tab:robustness}). The second most resistant approach, WC, is also close to our method conceptually, as it discourages the fine-tuned weights to deviate from pre-trained weights.

\begin{table}[ht]
\footnotesize
\setlength{\tabcolsep}{3pt}
\centering
\vspace{-0.4cm}
\begin{tabular}{lccccccc}
\toprule
Noise $\downarrow$     & STD++ &    DA &    WC & ReInit &  R3F &             \cnrt & \cprt \\
\midrule
$0\%$& 64.7 &               78.5 &                81.4 &  79.9 & 72.9 & {\bf 84.0} &               83.0 \\



        $5\%$ & 58.3 & 68.2 & 75.3 &  72.3 & 57.3 & {\bf 81.4} &   78.0 \\
        $10\%$ & 58.0 & 63.7 & 72.2 & 68.9 & 52.4 & {\bf 78.1} &              75.6 \\
        $20\%$ & 48.4 & 49.1 &              64.1 & 55.2 & 44.3 & 66.2 & {\bf 70.2} \\
        $30\%$ & 40.1 & 45.9 &              53.5 & 52.4 & 42.0 & 50.3 & {\bf 59.5} \\
\bottomrule
\end{tabular}
\caption{Mean performance over 13 datasets when training with noisy data. \cnrt: \concort, \cprt: \capcort. See  Appendix~\ref{sec:detailed_robustness_noise} for detailed results.  
}
\label{tab:robustness}
\end{table}

\subsection{Analyses on Few-shot Fine-tuning}
\label{subsection:fewshot}

To investigate our methods' robustness to small dataset sizes, we study \capcort and \concort performance in limited data settings (250/500/1000 training data points). 
We fix the same data subset across all models to avoid performance changes related to data variability. 

Since finetuning in few-shot setting is particularly prone to instability and the performance on a single dataset can distort the mean statistics for the entire collection, we use average rank as a more stable metric to compare different methods. A method's rank for a task corresponds to the position of the method in a list of all methods sorted by performance on that dataset. The minimal and best possible rank is $1$.  The {\em average rank} of a method is obtained by averaging ranks across all tasks. 

We observe in Table~\ref{tab:limited_data_results} that {\bf \concort is the most effective method in the few-shot setting measured in terms of the average rank}.  See Appendix~\ref{sec:small_datasets_full_results} for a detailed analysis.
\begin{table}[ht]
\footnotesize
\setlength{\tabcolsep}{3pt}
\centering
\begin{tabular}{ccccccccc}
\toprule
     \# samples $\downarrow$  &    STD++ &    DA &                  WC &              ReInit &   R3F &             \cnrt &             \cprt \\
\midrule
250 &  5.62 &                4.92 &                4.62 &                 3.00 &  4.04 &   {\bf 2.50} &                3.31 \\

     
500 &  6.08 &                4.38 &                3.69 &                3.31 &                4.85 &  {\bf 2.77} &                2.92 \\

1000 &  5.69 &   4.00 &                3.62 &                3.54 &  4.62 &  {\bf 2.69} &  3.85 \\

\bottomrule
\end{tabular}
\caption{Average rank of different methods for few-shot learning. \cnrt: \concort, \cprt: \capcort. 
}\label{tab:limited_data_results}
\end{table}

Overall, we find that \capcort yields performance gain on large-scale datasets, whereas \concort is effective for few-sample fine-tuning (since newly introduced parameters in \capcort are undertrained when the training data is limited). For wall-time analysis, see  Appendix~\ref{sec:walltime}. For experiments on hyper-parameter optimization over learning rate, batch size and other hyper-parameters see Appendix~\ref{sec:hpo}. We observe that \cnrt/\cprt are best performing models for all the datasets even in this setting. We also observe that our technique applied to RoBERTa achieves similar gains; our method is model independent (see Appendix~\ref{sec:roberta} for details).

\section{Degradation of Representations: Analytical Perspective}
\label{sec:representation_collapse_measures}

Here we quantify the representation collapse. 

\subsection{Probing Representation Collapse}
We follow the setting of \citet{AghajanyanSGGZG20} for studying representation collapse with {\em probing experiments} as follows: (i) fine-tune model on a downstream task $A$, (ii) freeze the encoding layers and train the top linear layer for a different task $B$. Low performance in the second step implies representation collapse in the first step. To assess robustness of the proposed approach to representation collapse, we perform a series of probing experiments. In our experimetnts, we use four GLUE and four non-GLUE datasets in the first step and all datasets in the second step except the one used in the first step (Table \ref{tab:next_task_finetuning_all_datapoints}). 
 
\begin{table}[ht]
\footnotesize
\setlength{\tabcolsep}{3pt}
\centering
\setlength{\tabcolsep}{3pt}
\begin{tabular}{ccccccccc}
\toprule
Task A $\downarrow$  & \stdpp &    DA &                  WC & ReInit &   R3F &             \cnrt & \cprt \\
\midrule
QNLI  & 37.6 & 37.1 &               44.5 &  37.7 &               36.5 &   41.7 & { \bf 49.5 } \\
QQP  & 39.8 & 42.6 &               44.5 &  41.2 &  36.3 & { \bf 52.4 } &   44.1 \\
RTE  & 32.0 & 32.0 & 37.2 &               48.9 & 33.3 & { \bf 51.9 } &               42.0 \\
MNLI  &               36.3 &               40.6 &  41.3 & { \bf 52.5 } & 43.0 &               51.0 &               48.5 \\
AG  & 41.1 &  42.5 & 42.3 &               43.3 & 41.4 &               43.7 & { \bf 47.8 } \\
IMDB   &               45.2 &               44.0 &               43.3 &  42.0 &               44.5 &               47.9 & { \bf 48.4 } \\
SCIT  & 39.0 & { \bf 50.3 } &               46.2 &  44.6 & 34.0 &               47.8 &                48.8 \\
SCIC &               43.9 &              44.7 &               46.3 &  41.4 &               39.1 & { \bf 48.3 } &               48.1 \\
\midrule
Aver. & 39.4 & 41.7& 43.2& 44.0& 38.5& {\bf 48.1}&47.1\\
\bottomrule
\end{tabular}
\caption{Results of probing experiments to measure representation collapse (higher score is better). Model is fine-tuned for task A with different methods, then a new linear head is trained for the remaining 12 tasks and the mean performance is reported. Aver. is  average over different choices of task A. \cnrt is \concort, \cprt is \capcort. 
AG: AGNEWS-10k, SCIT: SCITAIL-10k, SCIC: SCICITE-10k, QNLI: QNLI-10k, QQP:QQP-10k, MNLI: MNLI-10k. }
\label{tab:next_task_finetuning_all_datapoints}
\end{table}

We observe that \capcort and \concort show high resistance to representation collapse, outperforming other approaches in 6/8 cases (Table~\ref{tab:next_task_finetuning_all_datapoints}). For instance, fine-tuning for QNLI-10k in the first step with \capcort results in a mean performance of $49.5$ in the second step, whereas the next best baseline results in a mean performance of $44.5$.

Note that auxiliary tasks considered here are used only to evaluate the degradation of representations. They are not available during finetuning. During fine-tuning stage, only one task dataset is available. Thus, we do not compare our methods to the rehearsal-based learning methods.

\subsection{Diversity of Fine-tuned Representations}

Probing experiments rely on the availability of extra fine-tuning tasks and, thus, are limited in the amount of information they can assess, requiring additional fine-tuning rounds. Here, we propose metrics that can reliably quantify the power of fine-tuned representations by capturing their geometric diversity. The intuition behind our metrics is the following: {\em if all representations lie in a small dimensional space such as a straight line or a single point, then they contain little information and are not expressive. But if representations are well spread out and span the entire representation space, then they possess high information capacity}).

\begin{figure}[ht]
    \centering
    \includegraphics[scale=0.80]{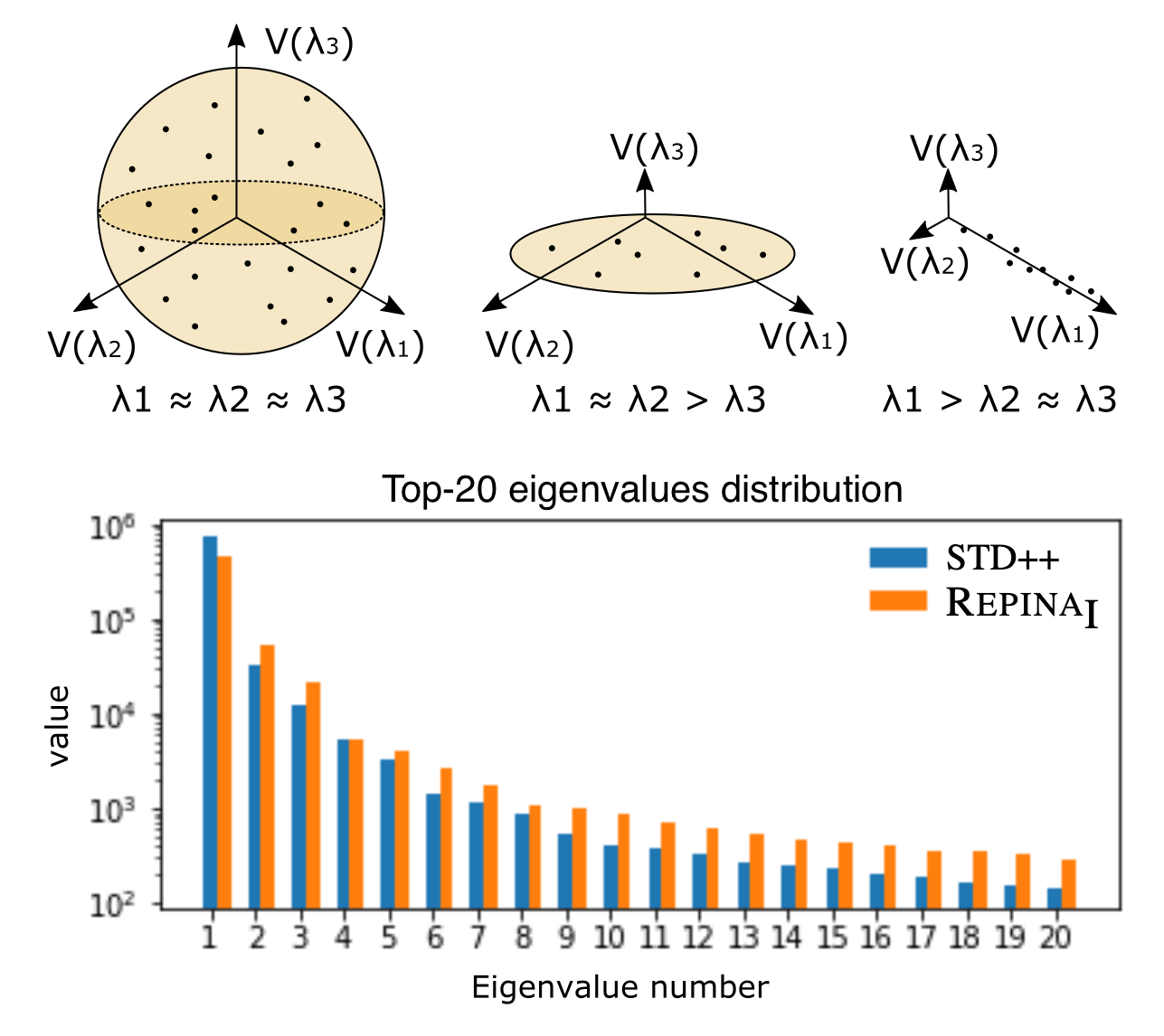}
    \caption{
    Top: 
    $\lambda_i$ and $V(\lambda_i)$ correspond to $i$th eigenvalue and its associated eigenvector after eigendecomposition of Gram matrix. Data from the left distribution is well spread out and spans all three dimensions, with all of its eigenvalues being similar. The right distribution
    shows all of the data collapsed along one eigenvector, hence one of the eigenvalues significantly exceeds two others. 
    Bottom: comparison of top-20 eigenvalues of \stdpp and \concort after fine-tuning on QQP with 250 points. Less skewed distribution of eigenvalues for \concort compared to \stdpp indicates a more spread out distribution of fine-tuned representations with \concort compared to \stdpp.
}
    \label{fig:eigen}
\end{figure}

We illustrate representation collapse metrics from the geometric perspective in Figure \ref{fig:eigen}. The top three plots show three different distributions of data points (representations). 
The left distribution spans all three dimensions, indicating the highest degree of data diversity. Since data points equally lie in all dimensions, all three eigenvectors ($V(\lambda_i)$'s) will be of equal importance and all three eigenvalues ($\lambda_i$'s) will be approximately equal. In contrast, the central distribution spans two axes, leading to a smaller third eigenvalue that corresponds to the "redundant" dimension. Right distribution has all the data points collapsed along one axis, resulting in one eigenvalue being substantially higher than the others. Overall, more uniform distribution of the eigenvalues corresponds to a better representation matrix diversity. In the bottom bar-plot we show distribution of the top-20 eigenvalues of the fine-tuned representations with \concort and \stdpp after training on QQP dataset with 250 points (Figure \ref{fig:eigen}). \concort preserved a closer to uniform eigenvalue distribution, while \stdpp results in representations with much higher first eigenvalue, indicating representation collapse. Thus, {\em \concort yields better representation matrix diversity and less representation collapse than \stdpp}.

Next, we formalize this intuition by defining a representation diversity metric based on the geometric diversity of sentence-level representations. 
\paragraph{Diversity Metrics:}
We compute the gram matrix $G$ for the representations  where $G_{i,j} = \langle f_{fin}(x_i),f_fin(x_j) \rangle$. From $G$ we obtain eigenvalues $\lambda_1\geq \dots \geq \lambda_d$. To measure diversity of representations, we use geometric mean (GM) and harmonic mean (HM) of the eigenvalues:
{
\begin{eqnarray*}
    &\firdivmeasure = \left(\Pi_{i=1}^d \lambda_i \right)^{1/d} = \text{Determinant}^{1/d}(G),\\
    &\secdivmeasure  = \left( \sum_{i=1}^d \frac{1}{\lambda_i}\right)^{-1} = \text{Trace}\left(G^{-1}\right)^{-1}
\end{eqnarray*}
}
These metrics attain a high value if the representations are well spread out and are zero or close to zero if all/most of the representations lie in a smaller dimension subspace. In contrast to the arithmetic mean, geometric and harmonic mean are not as sensitive to outliers. 

We observe that these metrics turn out to be always zero as representations typically lie in 20-dimensional space. Hence, we chose top-$k$ $\lambda_i$ values for $k=5,10,20$ where \firdivmeasure\space and \secdivmeasure\space are bounded away from $0$.

{
\begin{equation*}
    \firdivmeasure\text{-k} = \left(\Pi_{i=1}^k \lambda_i \right)^{\frac{1}{k}},\secdivmeasure\text{-k}  = \left( \sum_{i=1}^k \frac{1}{\lambda_i}\right)^{-1}
\end{equation*}
}
\begin{table}[ht]
\footnotesize
\setlength{\tabcolsep}{3pt}
\centering
\begin{tabular}{cccccccc}
\toprule
       Metric $\downarrow$  &  STD++ &                   DA &                   WC & ReInit &    R3F &              \cnrt & \cprt \\
        \midrule
\firdivmeasure-5 &  396 &               481 &               484 &  425 &                397 & { \bf 584 } &                463 \\
\firdivmeasure-10 &  92 &               118 &               118 &  90 &   93 & { \bf 134 } &    91 \\

\firdivmeasure-20 &               14 &               18 &               20 &              13 & 13 &  { \bf 22 } &   13 \\

\secdivmeasure-5 & 198 &               253 &               242 & 184 &               207 & { \bf 290 } &               217 \\

\secdivmeasure-10 &   38 &                53 &               47 &  37 &  38 &  { \bf 55 } &   32 \\

\secdivmeasure-20 &  3 &                4 &                5 &                3 &  3 &  { \bf 6 } &    3 \\
\bottomrule
\end{tabular}
\caption{Diversity of fine-tuned representations.
Mean value across all the 13 tasks is presented. \cnrt is \concort, \cprt is \capcort. \concort yields finetuned representations with maximum representation matrix diversity.
}
\label{tab:gm_hm_statistics_methods_comparison}
\end{table}
We compare \concort and \capcort to the existing baselines using \firdivmeasure-k and \secdivmeasure-k with $k=5,10,20$ (Table~\ref{tab:gm_hm_statistics_methods_comparison}). Low \firdivmeasure-k and \secdivmeasure-k indicates representation collapse, when  fine-tuned representations lie in a low-dimensional space. High \firdivmeasure-k and \secdivmeasure-k indicates that representations are well spread out and span a higher dimensional space. Table~\ref{tab:gm_hm_statistics_methods_comparison} shows that {\bf \concort results in the most diverse representations among all the baseline methods and incurs least representation collapse} (see Appendix~\ref{sec:secondary_metrics_detailed} for detailed results).  


\section{Conclusion}

In this paper, we propose a novel representation invariance regularizer targeted at avoiding representation degradation during finetuning. 
It has a knob that can control strength of regularization. We experiment with two choices of this knob, \concort and \capcort and show that they both achieve significant performance gain across 13 tasks, including few-shot and label noise settings, and improve generalization performance.  We also study the degradation of representations during fine-tuning, {\em representation collapse}, and propose new metrics to quantify it. Our methods reduce representation collapse and improve OOD robustness.
\section{Limitations}
We conduct extensive experiments in our paper and show that the proposed approaches lead to significant gains. However, we did not exhaust all avenues of investigation due to limited resources. 
Firstly, we could experiment with different choices of $\phi$ other than in \concort ($\phi$ is identity) and \capcort ($\phi$ is MLP). Other choices of $\phi$ may include deeper networks or transformer-based models, which could potentially improve performance even further.
Secondly, we investigated how representations from intermediate layers affect \concort performance, and observe major improvements with top layer representations. Similar experiments for \capcort may also yield further gain. 
Also, in \concort we could experiment with more choices of the representations layer (we tried 5th, 10th, 20th layer). 
Since lower layer representations are more computationally efficient to regularize (do not require full forward pass through the model), another interesting direction is finding a trade-off between computational efficiency of the regularizer and performance gains.

This study primarily focused on medium-sized models due to computational constraints. It is the outcome of extensive experimentation, which would have been impractical with limited computational resources. Although we have experimented with masked language models, we believe the findings apply to other architectures that follow similar principles. We anticipate that future research will provide more insight into these issues.

\section*{Ethical Considerations}
\textsc{REPINA} aims to improve performance and retain the quality of representations during fine-tuning. Practically, our method can help in suppressing potential biases of language models after fine-tuning on datasets that include biased content. \textsc{REPINA} can achieve this by reducing collapse of representations and preserving their pre-trained knowledge. All our experiments are based on publicly available datasets and, hence, there is no immediate concern about harmful content. 

\section*{Acknowledgments}
We thank the anonymous reviewers for their  constructive feedback. 
DK is supported by generous gifts from Johns Hopkins University, Allen Institute for AI and Amazon. AR is supported by Vector Institute for Artificial Intelligence and the University of Toronto. 

\clearpage
\bibliography{anthology,custom}
\bibliographystyle{acl_natbib}

\clearpage
{\large\textbf{Appendix}}

\section{Additional Related works}\label{sec:related_work_expanded}

Due to limited space in the main text, part of the related is below. We will reintroduce these to the main text upon having more space.

\paragraph{Domain shift between pre-training and finetuning data:} Even though pretrained models achieve high performance for a large number of NLP tasks, they tend to suffer if there is a significant domain shift between the pretraining data and finetuning data. Domain Adaptation bridges this gap by adapting  the model to the finetuning task domain. It can done by doing additional pre-training on task domain data if such data is available ~\citep{GururanganMSLBDS20} or algorithmically finding such data from general domain corpus if such a data is not available~\citep{MadanKZ21}. 

\paragraph{Domain shift between finetuning train data and evaluation data:} Domain Adaptation typically refers to the scenario where labeled train data is available in one domain and the evaluation is done for data in other domain. Techniques for addressing domain shift include model-centric techniques, data-centric techniques and hybrid techniques. Model-centric technique changes the model by changing the feature space, loss function or the structure of the model~\citep{BlitzerMP06,PanNSYC10,GaninUAGLLML16,BenRR20}. Data-centeric approaches involve pseudo-labeling~\cite{Abney07}, using auxiliary tasks~\cite{Phang18}, and data selection~\cite{MooreL10,WangULCS17}.  
{\em Mixout}~\citep{LeeCK19} is a variant of Dropout regularization that replaces dropped neurons with the pre-trained model neurons, thereby mixing pre-trained and fine-tuned parameters.

 \paragraph{Measures of representation:} ~\cite{AghajanyanSGGZG20} measures the quality of finetuned representations by fitting them on auxiliary tasks. CKA~\cite{KornblithNLH19} measures correspondences between representations from different network. ~\cite{WuBSDDG20} study the similarity of internal representation and attention of different trained models using some new similarity measures. 
 
 ~\cite{MerchantRPT20} also studies what happens during finetuning via probing classifiers and representation similarity analysis. It argues that finetuning does not necessarily incurs catastrophic forgetting. It analyze the effect for finetuning different tasks on the changes in representation.

 ~\cite{RongaliJRY20} show that rehearsal based learning can improve performance and perform better than Weight Consolidation. However, even though our method is inspired by multi-task learning and performs pseudo multi-task learning implicitly, we do not have access to any dataset additional to the single fine-tuning task. Thus, rehearsal based learning does not apply in our setting.

\paragraph{Parameter Efficient Finetuning:} Rather than storing a model for each of the finetuning task, some approaches try to keep most of the model parameters frozen and only tune a subset of parameters. ~\citet{RebuffiBV17} and ~\citet{PfeifferVGR20} insert adapter layers between the layers of pretrained model and keep the original parameters frozen. ~\citet{GuoRK20} keep the change in model parameters to be sparse. ~\cite{AghajanyanZG20} learns a small dimensional vector whose projection onto a large dimension space added to pretrained model parameter yields the finetuned model. ~\citet{BenRG21} show that finetuning only the bias parameters can also lead to competitive performance.

\paragraph{Multi-task learning:} 
In multi-task learning, we jointly finetune for many tasks where each task has a classification head but share the backbone with several other tasks~\citep{Caruana97,Ruder17,ZhangY17,LiuHCG19}.
This approach however requires access to a large amount of labeled data from unrelated tasks which is typically unavailable. Since our method focuses on the scenario when a single finetuning task data is available, we focus on comparing it against works of a similar nature and do not provide an extensive comparison with these works. It is likely that the clear difference between the methods makes them complementary, but exploring this is outside the scope of this paper.

\paragraph{Text-to-text finetuning:} Autoregressive models such as T5 and GPT-3 cast the finetuning in a text-to-text format~\cite{RaffelSRLNMZLL19,BrownMRSKDNSSA20}. They can work in the few-shot learning setting by framing the finetuning task as the pre-training task. Autoregressive models make it easier to sample text whereas masked Language models such as BERT and RoBERTa are restricted fill in the blanks.

\clearpage

\section{Theoretical Motivation and Connection to Pseudo Meta-learning}\label{sec:pseudo_meta_learning}

\paragraph{Idea:} We view the pre-trained model as a multi-task learner with an infinite number of pseudo tasks $T_1,T_2,\dots$. That is, for each $i$ there exists a linear layer that fits pre-trained representations to a pseudo task $T_i$. Our aim is to fine-tune the representations on a specific downstream task $B$ while preserving their ability to perform well on $T_1,T_2,\dots$; namely, there must exist a linear model on the fine-tuned representations for each pseudo task $T_i$. The linear classification head for $T_i$ does not have to be the same for the pre-trained and fine-tuned representations, but their output should be close. 
\begin{figure}[ht]
\includegraphics[width=0.49\textwidth]{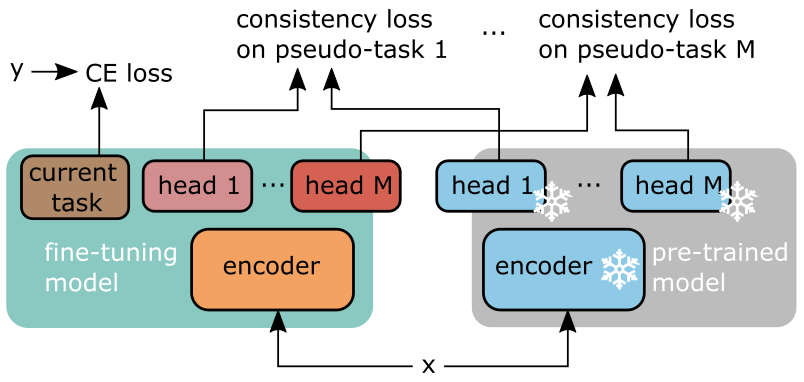}
\caption{Intuitive explanation of the proposed approaches from the multi-task learning perspective. The total loss consists of the cross-entropy loss for the fine-tuning task and the consistency losses for the pseudo-tasks. The pre-trained model is non-trainable (frozen).}

\label{fig:theory_illustration}
\end{figure}

Let the training samples be $x_1,\dots,x_N$ for the fine-tuning task $B$ and $\zprej,\zfinj \in \rR^d$ be the representations (output of the encoder layer) for the pre-trained model and the model being fine-tuned. Let $\fF$ be a family of functions operating on representations such that each function signifies a task, $T_i$. We can formalize our objective as follows: {\em For any function $g\in \fF$ on pre-trained representations, there must exists a corresponding $h \in \fF$ on fine-tuned representations giving the same output; $\forall g \in \fF, \exists h \in \fF \text{ s.t. }  g\circ z_{pre} = h \circ z_{fin}$ }

During fine-tuning, we do not expect an exact agreement and allow representations to lose some expressive power. Hence, we relax the constraint and consider the representation loss error.\footnote{If $T_i$'s were actual pre-training tasks and the data was is available, we would compute the loss on the input of $T_i$. In absence of that, we approximate it by loss on (unlabeled) input of the given fine-tuning task.} 

\[\text{For $ g \in \fF$, }\mathcal{L}_g = \min_{h \in \fF} \sum_{j=1}^N \text{loss}(g(\zprej),{ } h(\zfinj))\]
If $g$ comes from a distribution $\dD$ over $\fF$, then our representation loss is $E_{g \sim \dD} \left[\mathcal{L}_g\right]$. Here, we consider $\fF$ to be the set of regression tasks\footnote{If we assume the tasks to be classification, then the linear layer is followed by a softmax layer. However, for simplicity we assume the pre-training to be done on regression tasks as this yields closed from expressions and yields good results.}, which are characterized by vectors in $\{u \in \rR^d\}$ and tasks to be sampled from a standard Gaussian distribution with mean ${\bf 0}$ and unit variance. We consider $loss$ to be the natural $\ell_2$ loss function. 
\begin{align*}
\mathcal{L} = 
&E_{u \sim \nN({\bf 0},\iI_d)} \left[\min_{v \in \rR^d}\sum_{j=1}^N(u^T \zprej - v^T \zfinj)^2\right]
\end{align*}

The inner minimization has a closed form solution 
$v = (Z_{fin}^T Z_{fin})^{\dagger} Z_{fin}^T Z_{pre} u$,  
and the resulting expectation can be reduced to get: 
$$\mathcal{L}= \left\|\left(Z_{fin} (Z_{fin}^T Z_{fin})^{\dagger} Z_{fin}^T - I_n\right) Z_{pre}\right\|_2^2.$$
where $Z_{pre}\in \rR^{N \times d}$ matrix has $j$-th row  $\zprej$ and $Z_{fin} \in \rR^{N \times d}$ matrix has $j$-th row $\zfinj$. $||\cdot||_2$ for a vector denote the $\ell_2$-norm and for a matrix denote the Frobenius norm. $A^{\dagger}$ denote the pseudo-inverse of a symmetric matrix $A$. The derivation of $v$ and reduction of expectation can be found in Theorem~\ref{thm:1} in Appendix~\ref{sec:infinite_continual_learning_derivations}.
Loss function $\hat{L}$ is not easily decomposed into mini-batches, making it challenging to optimize directly.
%
We find an equivalent optimization problem whose objective is decomposable into mini-batches and whose optimum is equivalent to the representation loss $\mathcal{L}$: 
\begin{align*}
\hat{\mathcal{L}}
    &= \min_{W \in \rR^{d \times d}} \sum_{j=1}^N \left\|\zprej - W\zfinj \right\|_2^2
\end{align*}


We can minimize $\hat{\mathcal{L}}$ for $W$ along with the fine-tuning objective. 
Derivation of the above equivalence can be found in Theorem~\ref{thm:2} in Appendix~\ref{sec:infinite_continual_learning_derivations}. We can interpret the above loss as follows: {\em There exists a linear function ($\phi_W:x\rightarrow Wx$) which operated on fine-tuned representations results in pre-trained representations.} We can generalize it to include a class of functions $\Phi$: 
\begin{equation}\label{eq:rep_loss}
    \hat{\mathcal{L}}=\min_{\phi \in \Phi} \sum_{j=1}^N \left\|\zprej - \phi(\zfinj)\right\|_2^2
\end{equation}
This corresponds to the aggregate loss for pseudo-tasks $T_i$'s, if instead of using a linear head for pseudo tasks on fine-tuned representations, we use a function $\phi \in \Phi$ followed by $u_i$. Thus, $\Phi$ defines the strength of the regularizer. A singleton $\Phi$ containing an identity function enforces the use of the same linear head $u_i$ for task $T_i$ on both pre-trained and fine-tuned representations.
This results in the strongest regularizer which keeps fine-tuned representations close to the pre-trained representations. On the other hand, if $\Phi$ is a set of very deep neural networks, then we allow a deep neural network ($+u_i$) to fit fine-tuned representations for task $T_i$. Such a neural network will almost always exist even if the fine-tuned representations have degraded significantly. Thus, it is a weak regularizer and puts mild constraints on the change of the structure of representations.

Overall, this section can be summarized as follows: (i) $\hat{\mathcal{L}}$ is an aggregate error in fitting fine-tuned representations to pseudo-pre-training tasks $T_i$'s. 
(ii) $\Phi$ controls the amount of structural changes in representations allowed during fine-tuning. 

\subsection{Detailed Derivations}\label{sec:infinite_continual_learning_derivations}

\begin{lemma}\label{lem:linear_regression_loss}
$min_{v \in \rR^d} \sum_{j=1}^N (y_j - v^T b_j)^2 =||y - B (B^T B)^\dagger B^T y||_2^2$ where $y_j$ is the $j$-th entry of $y$. 
\end{lemma}
\begin{proof}
Let the loss function be 
\[ \mathcal{L} = \sum_{j=1}^N (y_j - v^T b_j)^2\]
$\mathcal{L}$ is a smooth function with minimizer $v^\star$. Hence, minimum is achieved at a local minimum. Thus, 
\begin{align*}
    \frac{\delta}{\delta v} \mathcal{L}|_{v=v^\star} &= {\bf 0}\\
    -2 \sum_{j=1}^N b_j(y_j - (v^\star)^T b_j) & = {\bf 0}\\
    -2 \sum_{j=1}^N b_j y_j - b_j b_j^T v^\star &= {\bf 0}\\
    \left(\sum_{j=1}^N b_j \right)y_j & =\left( \sum_{j=1}^N b_j b_j^T\right)v^{\star}\\
    \left( \sum_{j=1}^N b_j b_j^T\right)^{\dagger} \left(\sum_{j=1}^N b_j \right)y_j& = v^{\star}
\end{align*}
where $X^{\star}$ is the pseudo inverse which is equal to the inverse if $X$ is invertible. Else it spans only the space spanned by $X$. Note that $\sum_{j=1}^N b_j b_j^T = B^TB$ and $\sum_{j=1}^N b_j y_j = B^T y$. So, $v^\star = (B^T B)^{\dagger} B^T y$. Least square error can be written in terms of vector form to get
\[min_{v \in \rR^d} \sum_{j=1}^N (y_j - v^T b_j)^2 = \min_{v \in \rR^d} \left|\left|y - Bv\right|\right|_2^2\]
where $||\cdot||_2^2$ for a vector denote the $\ell_2$ norm squared.
Substituting $v^*$ we get
$min_{v \in \rR^d} \sum_{j=1}^N (y_j - v^T b_j)^2 =\left|\left|y - B (B^T B)^\dagger B^T y\right|\right|_2^2$
\end{proof}

\begin{theorem}\label{thm:1}
We show that
$E_{u \sim \nN({\bf 0},\iI_d)} \left[\min_{v \in \rR^d}\sum_{j=1}^N(u^T \zprej - v^T \zfinj)^2\right]$ $=\left\|\left(Z_{fin} (Z_{fin}^T Z_{fin})^{\dagger} Z_{fin}^T - I_n\right) Z_{pre}\right\|_2^2.$
\end{theorem}
\begin{proof}
To simplify notation, we use $a_j = z_{pre}^j$,$b_j = z_{fin}^j$,$ B=Z_{fin} \in \rR^{N \times d}$ matrix has $j$-th row $b_j$, $A = Z_{pre} \in \rR^{N \times d}$ matrix has $j$-th row $a_j$ and $X^{\dagger}$ is the pseudo-inverse of $X$.

Let \[W = E_{u \sim \nN({\bf 0},\iI_d)} \left[\min_{v \in \rR^d}\sum_{j=1}^N(u^T \zprej - v^T \zfinj)^2\right]\] From Lemma~\ref{lem:linear_regression_loss}, we get 
\begin{align*}
W &= E_{u \sim \nN({\bf 0},\iI_d)}\left|\left|Au - B (B^T B)^\dagger B^T A u\right|\right|_2^2\\
&= E_{u \sim \nN({\bf 0},\iI_d)}\left|\left|(A - B (B^T B)^\dagger B^T A) u\right|\right|_2^2
\end{align*}
\begin{lemma}\label{lem:gaussian_expectation}
For any matrix $M$, $\rR^{d\times d}$. $E_{u \sim \nN({\bf 0},\iI_d)}[ \left|\left|Mu\right|\right|_2^2] = \left|\left|M\right|\right|_2^2$ where $\left|\left|M\right|\right|_2^2$ is the forbenius norm of the matrix $M$.
\end{lemma}
\begin{proof}
Let the $i,j$-th entry of $M$ be $m_{i,j}$ and the $j$-th entry in $u$ be $u_j$. Then, $\left| \left|Mu\right|\right|_2^2 = \sum_{i=1}^d (\sum_{j=1}^d m_{i,j} u_j)^2= \sum_{i=1}^d \sum_{j=1}^d \sum_{k=1}^d m_{i,j} m_{i,k} u_j u_k$.

\[ E\left[\left|\left|Mu\right|\right\|_2^2\right] = \sum_{i=1}^d \sum_{j=1}^d \sum_{k=1}^d m_{i,j} m_{i,k} E[u_j u_k]\]
Since $u$ is a gaussian random variable with mean $0$ and covariance matrix $I_d$, we have $E[u_ju_k]=0$ for $j\neq k$ and $E[u_i^2] = 1$ for all $i \in [d]$. 
Thus,
\[E\left[\left|\left|Mu\right|\right\|_2^2\right] = \sum_{i=1}^d \sum_{j=1}^d m_{i,j}^2 = ||M||_2^2\]
\end{proof}
Substituting equality from Lemma~\ref{lem:gaussian_expectation} to $W$, we get
\[ W = \left|\left|A - B (B^T B)^\dagger B^T A\right|\right|_2^2\]
Using $||-M||_2^2 = ||M||_2^2$ and substituting back $A=Z_{pre}$ and $B = Z_{fin}$, we get
$E_{u \sim \nN({\bf 0},\iI_d)} \left[\min_{v \in \rR^d}\sum_{j=1}^N(u^T \zprej - v^T \zfinj)^2\right]$ $=\left\|\left(Z_{fin} (Z_{fin}^T Z_{fin})^{\dagger} Z_{fin}^T - I_n\right) Z_{pre}\right\|_2^2.$
\end{proof}

\begin{theorem}\label{thm:2}
\begin{align*}
& \left\|\left(Z_{fin} (Z_{fin}^T Z_{fin})^{\dagger} Z_{fin}^T - I_n\right) Z_{pre}\right\|_2^2\\
    &= \min_{W \in \rR^{d \times d}} \sum_{j=1}^N \left\|\zprej - W\zfinj \right\|_2^2
\end{align*}
\end{theorem}
\begin{proof}
To simplify notation, we use $a_j = z_{pre}^j$,$b_j = z_{fin}^j$,$ B=Z_{fin} \in \rR^{N \times d}$ matrix has $j$-th row $b_j$, $A = Z_{pre} \in \rR^{N \times d}$ matrix has $j$-th row $a_j$ and $X^{\dagger}$ is the pseudo-inverse of $X$. Let $W = \rR^{d\times d}$ have $i$-th row $w_i$. We need to compute
\begin{align*}
L &=     \min_{W \in \rR^{d \times d}} \sum_{j=1}^N \left\|a_j - Wb_j \right\|_2^2\\
&= \min_{w_1,\dots,w_d\in \rR^{d}} \sum_{j=1}^N \sum_{i=1}^d (a_{j,i} - w_i^T b_j)^2\\
& = \sum_{i=1}^d \min_{w_i \in \rR^d} \sum_{j=1}^N (a_{j,i} - w_i^T b_j)^2
\end{align*}
where $a_{j,i}$ is the $i$-th entry of $a_j$. Applying Lemma~\ref{lem:linear_regression_loss}, we get 
\begin{align}\label{eq:last}
    L & = \sum_{i=1}^d \left\|(I_n - B (B^T B)^\dagger B^T)c_i \right\|^2
\end{align}
where $c_i$ is the $i$-th column of $A$ ($j$-th entry of $c_i$ is $a_{j,i}$). 
\begin{lemma}\label{lem:matrix_vector_norm_temp}
For a matrix $M \in \rR^{N \times N}$ and a set of vectors $v_1,\dots,v_k \in \rR^N$, 
\[ \sum_{i=1}^k \left\|Mv_i\right\|^2 = ||MV||_2^2\]
where $V\in \rR^{N\times k}$ is the matrix with columns $v_1,\dots,v_k$.
\end{lemma}
\begin{proof}
Let $j$-th row of $M$ be $m_j$. Then, 
\[\sum_{i=1}^k \left\|Mv_i\right\|^2 = \sum_{i=1}^k\sum_{j=1}^N (m_j^T v)^2\]
For $j \in [N], i \in [k]$, $(j,i)$-the entry of $MV$ is $ m_j^T v_i$. Thus,
\[ \left\|MV\right\|^2 = \sum_{j=1}^N \sum_{i=1}^k (m_j^T v_i)^2\]
Combining the two equalities, we get 
\[ \sum_{i=1}^k \left\|Mv_i\right\|^2 = ||MV||_2^2\]
\end{proof}
Applying Lemma~\ref{lem:matrix_vector_norm_temp}in eq~\ref{eq:last}, we get

\[ L = \left\|(I_n - B (B^T B)^\dagger B^T)A \right\|^2\]
This finishes the proof of theorem.
\end{proof}

\clearpage

\section{Baselines - detailed}\label{sec:baselines_detailed}


\paragraph{Weight Consolidation:}~\cite{KirkpatrickPRVDRMQRG17,ChenHCCLY20} Let $\pP$ be the set of all model parameters and $\bB$ be the subset of the bias parameters (affine component in the linear transformations) 
of the model. For a parameter $\theta_i$, let $\theta_{i}^{pre}$ be the pre-trained value and $\theta_{i}^{fin}$ be the value during the finetuning process. Then, the regularization loss is.
\[ \mathcal{L}_{WC} = \sum_{i\in \pP \setminus \bB} ||\theta_{i}^{fin}-\theta_{i}^{pre}||_2^2 \]

\paragraph{Local smoothness inducing regularization R3F} ~\citep{AghajanyanSGGZG20} For a classification problem, let $f$ be the probability prediction function corresponding to model being finetuned. It's input is the input to the first BERT encoder layer (output of token embedding layer). Let $x_1,\dots,x_N$ be the outputs of the token embedding layer for the inputs of the finetuning task. For $i\in [N]$, let $\epsilon_{\delta, i}$ be a Gaussian noise term with mean ${\bf 0}$ and covariance matrix $\delta I$. We set $\delta=1e-5$. Then, the regularization loss is
\begin{align*} \mathcal{L}_{R3F} &= \sum_{i=1}^N KL(f(x_i)\mid\mid f(x_i+\epsilon_{\delta, i})) \\ &\hspace{3em}+ KL(f(x_i)+\epsilon_{\delta, i})\mid\mid f(x_i))
\end{align*}

\paragraph{Data Augmentation:} Let $x_1,\dots,x_N$ be the outputs of the token embedding layer for the inputs of the finetuning task and $y_1,\dots,y_N$ be their associated labels. In data augmentation, we add noise to the data during the training process.
 \[ \mathcal{L}_{total} = \sum_{i=1}^N \mathcal{L}_{CE}(f(x_i),y_i) + \lambda \cdot \mathcal{L}_{CE}(f(x_i+\epsilon_{\delta,i}),y_i)\]
 where $\lL_{CE}$ is the cross entropy loss, $f$ is the prediction function as per the model and $\epsilon_{\delta,i}$ is a Gaussian noise with mean ${\bf 0}$ and co-variance matrix $\delta \iI$ added to $x_i$. We set $\delta=1e-5$.

Table~\ref{tab:reg_coefficients} show the regularization coefficients used for each method.

\begin{table}[h]
    \centering
    \small
    \begin{tabular}{cc}
    \toprule
         Method & Regularization coefficient \\
    \midrule
       \concort  & 0.01, 0.05, 0.1, 0.5\\
       \capcort & 0.01, 0.05, 0.1, 0.5\\
       DA & 0.05, 0.1, 0.2, 0.4, 0.8\\
       R3F & 0.1, 0.5, 1, 5\\
       WC &0.01, 0.05, 0.1, 0.5 \\
       \bottomrule
    \end{tabular}
    \caption{Regularization coefficient for different methods.}
    \label{tab:reg_coefficients}
\end{table}


\clearpage
\section{Experiment Set up details}\label{sec:experiment_set_up_appendix}
Our implementation is based on the HuggingFace library.


\subsection{Experimental Setup}
\label{appendix:experiment}
To avoid the excessively high computational cost of fine-tuning on large-scale datasets, we limited their full training sets to $10,000$ data points (marked with a suffix -10k in Table~\ref{tab:datasets}). For few-sample experiments, we fixed the same data subset across all models to avoid performance changes related to data variability. Since test set labels are unavailable, we use development set to report the performance.

\paragraph{Batch Size:} Different methods have different memory requirement. For instance, R3F has the highest footprint which limits the batch size as we can not process too many inputs at the same time. Table~\ref{tab:batch_size} shows the batch size used for each dataset in our experiments.
\begin{table}[h]
\footnotesize
\centering
\setlength{\tabcolsep}{1pt}
\begin{tabular}{cc}
\toprule
    Task &  Batch size\\
\midrule
    COLA &   4\\
     RTE &   1\\
    SST & 4\\
    MNLI-10k & 1\\
    MRPC &   4\\
    QQP-10k & 1\\
    QNLI-10k & 1\\
    \bottomrule
\end{tabular}
\begin{tabular}{cc}
\toprule
    Task &  Batch size\\
\midrule
     CHEMPROT &   1\\
     SCICITE &   2\\
     SCITAIL-10k &  1\\
  AGNEWS-10k & 2\\
    YELP-10k & 1\\
    IMDB-10k &  1\\
    &\\
    \bottomrule
\end{tabular}
\caption{Batch size used in our experiments}
\label{tab:batch_size}
\end{table}

\newpage
\subsection{Filtering Failed Runs}
\label{appendix:failed:runs}
For most of the datasets experimented here, available test data split is unlabeled. Thus, we use the validation data split to report performance. It has been observed that different finetuning runs can result in very different finetuned model performance~\cite{MosbachAK20}. Thus, reporting max test run performance does not truly reflect the effectiveness of the finetuning process and the maximum test run performance across different random seeds can be substantially larger than the mean. So, in our experiments we do not use the val data (on which we report performance) to select the run or any hyperparameter. To select optimal hyperparameters such as regularization coefficient etc., we use a subset of original train data split which is not used for training. Such a data is available as we are typically finetuning with a subset of original train data. Table~\ref{tab:failed_threshold} shows the threshold for failed run for each task.
\begin{table}[th]
\footnotesize
\centering
\setlength{\tabcolsep}{1pt}
\begin{tabular}{cc}
\toprule
    Task &  Threshold\\
\midrule
    COLA &   0.00\\
     RTE &   53.70\\
    SST & 54.00\\
    MNLI-10k & 30.00\\
    MRPC &   81.22\\
    QQP-10k & 0.00\\
    QNLI-10k & 50.53\\
    \bottomrule
\end{tabular}
\begin{tabular}{cc}
\toprule
    Task &  Threshold\\
\midrule
     CHEMPROT &   34.45\\
     SCICITE &   24.66\\
     SCITAIL-10k &  60.38\\
  AGNEWS-10k & 10.44\\
    YELP-10k & 52.80\\
    IMDB-10k &  33.94\\
    &\\
    \bottomrule
\end{tabular}
\caption{Failed run threshold}
\label{tab:failed_threshold}
\end{table}

\clearpage

\clearpage
\section{\capcort - Effect of MLP depth and missing details}\label{sec:capcort_mlp_depth}

\paragraph{Missing Details:} We use tanh activation in MLP with learning rate same as the rest of the network. Parameters of MLP are optimized alongside the language model. We use Glorot uniform initializer to initialize the parameters of MLP~\cite{GlorotB10}. Bias parameters are initialized to zero. 

Table~\ref{tab:capcort_layer_by_layer_all} 
shows that \capcort is resistant to the number of MLP layers chosen. When training with all datapoints, performance is typically within a percentage of each other. 

\begin{table}[ht]
\centering
\small
\begin{tabular}{lcc}
\toprule
       Task $\downarrow$ / Num. layers $\rightarrow$ &           1 &           2\\
\midrule
    IMDB-10k &               93.43 & {\bf 93.87} \\

        MRPC & {\bf 91.19} &               91.12 \\

    SCICITE &               82.55 & {\bf 83.15} \\

        COLA &               61.86 & {\bf 62.47} \\

        SST &               93.03 & {\bf 93.19} \\

    MNLI-10k &               65.18 &               {\bf 65.22} \\

  AGNEWS-10k &                91.8 & {\bf 91.93} \\

    CHEMPROT &               82.48 &               {\bf 83.67} \\

    QNLI-10k &               87.52 & {\bf 87.61} \\

 SCITAIL-10k &               94.06 & {\bf 94.75} \\

    YELP-10k &               95.66 & {\bf 95.96} \\

         RTE &               72.92 & {\bf 74.37} \\

     QQP-10k &               78.22 &  {\bf 79.3} \\
             Mean &               83.83 &               84.35 \\
\bottomrule
\end{tabular}
\caption{Performance of \capcort with different number of MLP layers.}\label{tab:capcort_layer_by_layer_all}
\end{table}











\clearpage
\section{\concort - Which layer to regularize?}\label{sec:concort_layer_by_layer_analysis}

Table~\ref{tab:concort_vs_concort_top_250},~\ref{tab:concort_vs_concort_top_500} and~\ref{tab:concort_vs_concort_top_1000} compares the result between regularizing the top layer vs regularizing the intermediate layer in \concort. We observe that \concort consistently outperform when regularizing the intermediate layer.

\begin{table}[ht]
\centering
\small
\begin{tabular}{cccc}
\toprule
       Tasks &               STD++ &                  Top &             Interm. \\
\midrule
\multicolumn{4}{c}{250 Training Datapoints}\\
\midrule
        QNLI &               75.11 &               76.12 & {\bf 78.82} \\

        MNLI &                37.7 & {\bf 38.67} &               38.35 \\

      AGNEWS &               88.08 &               87.78 & {\bf 88.53} \\

        IMDB &               86.11 &               87.44 & {\bf 90.38} \\

        SST &               88.41 &               88.58 & {\bf 89.29} \\

        COLA &               41.57 & {\bf 44.26} &               43.98 \\

    CHEMPROT &               55.22 &               59.53 & {\bf 63.28} \\

        MRPC &               84.43 &               83.91 & {\bf 84.65} \\

     SCITAIL &               82.31 &               85.03 &  {\bf 88.9} \\

    SCICITE & {\bf 76.86} &               75.64 &               76.11 \\

         RTE &               59.13 & {\bf 61.13} &               60.83 \\

        YELP &               92.51 &               92.23 & {\bf 93.13} \\

         QQP &               68.28 &               67.05 & {\bf 68.76} \\
\midrule
        Mean &               71.98 &               72.88 & {\bf 74.23} \\

Average Rank &                2.54 &                2.15 &  {\bf 1.31} \\
\bottomrule
\end{tabular}
\caption{Performance for \concort-intermediate vs \concort-top.}\label{tab:concort_vs_concort_top_250}
\end{table}
\begin{table}[ht]
\small
\centering
\begin{tabular}{cccc}
\toprule
       Tasks &               STD++ &                  Top &             Interm. \\

\midrule
\multicolumn{4}{c}{500 Training Datapoints}\\
\midrule
        QNLI &               80.86 &  80.7 & {\bf 82.56} \\

        MNLI &               41.74 & 41.07 & {\bf 49.07} \\

      AGNEWS &               89.09 & 89.34 & {\bf 89.53} \\

        IMDB &               83.65 & 89.77 & {\bf 91.32} \\

        SST &               89.54 & 89.49 & {\bf 89.68} \\

        COLA &               45.29 & 44.22 & {\bf 46.95} \\

    CHEMPROT &               65.59 &  61.0 & {\bf 73.47} \\

        MRPC &               84.44 & 84.54 & {\bf 85.26} \\

     SCITAIL &               85.11 & 88.97 & {\bf 90.07} \\

    SCICITE &               78.84 & 79.36 & {\bf 79.45} \\

         RTE &               61.01 & 59.39 & {\bf 62.45} \\

        YELP & {\bf 93.32} & 89.02 &                93.2 \\

         QQP &               61.12 &  69.5 & {\bf 70.97} \\
\midrule
        Mean &               73.81 & 74.34 & {\bf 77.23} \\

Average Rank &                2.38 &  2.54 &  {\bf 1.08} \\
\bottomrule
\end{tabular}
\caption{Performance for \concort-intermediate vs \concort-top.}\label{tab:concort_vs_concort_top_500}
\end{table}
\begin{table}[ht]
\small
\centering
\begin{tabular}{cccc}
\toprule
       Tasks &               STD++ &                  Top &             Intermediate \\

\midrule

\multicolumn{4}{c}{1000 Training Datapoints}\\
\midrule
        QNLI &               81.73 &               69.19 & {\bf 83.67} \\

        MNLI &               50.32 &               49.84 &  {\bf 55.5} \\

      AGNEWS &               89.29 &               89.45 & {\bf 89.74} \\

        IMDB &                90.9 &               72.39 & {\bf 91.35} \\

        SST & {\bf 90.69} &               90.12 &               90.32 \\

        COLA &               49.39 &               49.22 & {\bf 51.05} \\

    CHEMPROT &               75.64 &                74.6 & {\bf 79.64} \\

        MRPC & {\bf 87.82} &               86.93 &               87.54 \\

     SCITAIL &               80.31 &               90.08 & {\bf 91.51} \\

    SCICITE &               80.85 & {\bf 81.16} &               80.25 \\

         RTE &               63.63 &                63.9 & {\bf 67.06} \\

        YELP &                94.3 &               84.45 & {\bf 94.78} \\

         QQP &               70.58 &               67.22 & {\bf 73.04} \\
\midrule
        Mean &               77.34 &                74.5 & {\bf 79.65} \\

Average Rank &                2.08 &                2.62 &  {\bf 1.31} \\
\bottomrule
\end{tabular}
\caption{Performance for \concort-intermediate vs \concort-top.}\label{tab:concort_vs_concort_top_1000}
\end{table}

Table~\ref{tab:concort_layer_by_layer_250},~\ref{tab:concort_layer_by_layer_500}, ~\ref{tab:concort_layer_by_layer_1000} and ~\ref{tab:concort_layer_by_layer_all} show the the result for \concort with representations chosen from 5th, 10th or 20th layer of encoder. Note that 5th layer is the closest to the input and doesn't account for token embedding layer. We note that all three choices are performing roughly equally well. Mean performance is typically less than a percentage point from each other. If one were to use a single layer, one can use 5th for low-data case and 10th or 20th for large dataset case. 

\begin{table}[ht]
\small
\centering
\begin{tabular}{ccccc}
\toprule
Tasks & STD++ &          5 &          10 &           20\\
\midrule
\multicolumn{5}{c}{250 Training datapoints}\\
\midrule

        COLA & 41.57 &               41.23 & {\bf 43.98} &               41.94 \\

        QNLI & 75.11 & {\bf 78.82} &                77.9 &               75.28 \\

        MRPC & 84.43 &               83.99 &               84.65 & {\bf 84.81} \\

        SST & 88.41 &               88.42 &               89.29 & {\bf 89.45} \\

     SCITAIL & 82.31 & {\bf 89.48} &                88.9 &               88.07 \\

        YELP & 92.51 & {\bf 93.13} &               92.68 &               92.63 \\

      AGNEWS & 88.08 &               88.53 & {\bf 88.59} &               87.86 \\

         RTE & 59.13 &               60.83 &               59.78 & {\bf 61.23} \\

        MNLI &  37.7 &               38.35 & {\bf 41.16} &               39.81 \\

         QQP & 68.28 &               66.66 &               67.78 & {\bf 68.76} \\

        IMDB & 86.11 &                88.9 & {\bf 90.57} &               90.38 \\

    CHEMPROT & 55.22 &                58.0 &               60.53 & {\bf 63.28} \\

    SCICITE & 76.86 &               76.35 & {\bf 78.95} &               76.11 \\
\midrule
        Mean & 71.98 &               73.28 &               74.21 &               73.82 \\
\bottomrule
\end{tabular}
\caption{Effect of embedding layer to be regularized in \concort.}
\label{tab:concort_layer_by_layer_250}
\end{table}

\begin{table}[ht]
\small
\centering
\begin{tabular}{ccccc}
\toprule
Tasks & STD++ &          5 &          10 &           20\\
\midrule
\multicolumn{5}{c}{500 Training datapoints}\\
\midrule
        COLA &               45.29 & {\bf 46.95} &               44.58 &               44.34 \\

        QNLI &               80.86 & {\bf 82.56} &               81.38 &                80.7 \\

        MRPC &               84.44 & {\bf 85.26} &               84.78 &               84.31 \\

        SST &               89.54 & {\bf 89.68} &               89.24 &               89.66 \\

     SCITAIL &               85.11 & {\bf 90.36} &               90.07 &                89.2 \\

        YELP & {\bf 93.32} &               92.88 &                93.2 &                93.2 \\

      AGNEWS &               89.09 & {\bf 89.53} &               89.19 &               88.99 \\

         RTE &               61.01 &               62.45 & {\bf 63.63} &               61.52 \\

        MNLI &               41.74 &               44.61 & {\bf 49.07} &               44.57 \\

         QQP &               61.12 &  {\bf 72.6} &               71.79 &               70.97 \\

        IMDB &               83.65 &               90.84 &               90.31 & {\bf 91.32} \\

    CHEMPROT &               65.59 &               73.34 & {\bf 73.47} &               71.98 \\

    SCICITE &               78.84 &               79.39 &               78.76 & {\bf 79.45} \\
\midrule
        Mean &               73.81 &               76.96 &               76.88 &               76.17 \\
\bottomrule
\end{tabular}
\caption{Effect of embedding layer to be regularized in \concort.}
\label{tab:concort_layer_by_layer_500}
\end{table}
\begin{table}[ht]
\small
\centering
\begin{tabular}{ccccc}
\toprule
Tasks & STD++ &          5 &          10 &           20\\
\midrule
\multicolumn{5}{c}{1000 Training Datapoints}\\
\midrule

        COLA &               49.39 & {\bf 51.05} &               48.16 &               47.74 \\

        QNLI &               81.73 &               83.11 & {\bf 83.67} &               83.33 \\

        MRPC &               87.82 &               87.54 & {\bf 87.84} &               87.08 \\

        SST &               90.69 & {\bf 90.75} &               90.32 &               90.29 \\

     SCITAIL &               80.31 &               91.09 &               91.51 & {\bf 91.73} \\

        YELP &                94.3 & {\bf 94.78} &               94.39 &               94.32 \\

      AGNEWS &               89.29 &               89.47 &               89.74 & {\bf 89.84} \\

         RTE &               63.63 & {\bf 67.06} &                63.1 &               66.28 \\

        MNLI &               50.32 &                55.5 & {\bf 55.88} &                53.6 \\

         QQP &               70.58 &               70.86 &               72.21 & {\bf 73.04} \\

        IMDB &                90.9 &                90.9 &               91.35 & {\bf 91.47} \\

    CHEMPROT &               75.64 &               79.64 & {\bf 79.66} &               78.72 \\

    SCICITE & {\bf 80.85} &               80.83 &               80.35 &               80.25 \\
\midrule
        Mean &               77.34 &               79.43 &               79.09 &               79.05 \\

\bottomrule
\end{tabular}
\caption{Effect of embedding layer to be regularized in \concort.}
\label{tab:concort_layer_by_layer_1000}
\end{table}
\begin{table}[ht]
\small
\centering
\begin{tabular}{ccccc}
\toprule
Tasks & STD++ &          5 &          10 & 20\\
\midrule
\multicolumn{5}{c}{Full Datasets}\\
\midrule
        COLA & 59.69 & {\bf 62.34} &               61.65 &               58.45 \\

        MRPC & 86.84 &               90.01 & {\bf 91.49} &               89.66 \\

    IMDB-10k &  93.2 &               93.34 &                93.5 & {\bf 93.96} \\

 SCITAIL-10k & 71.61 &               93.71 & {\bf 93.74} &               93.63 \\

         RTE & 70.76 &               68.35 &               69.75 & {\bf 71.41} \\

    QNLI-10k & 87.23 & {\bf 87.52} &               87.46 &               87.37 \\

    YELP-10k & 95.33 &               95.62 &               95.78 & {\bf 96.01} \\

  AGNEWS-10k & 91.67 &               91.83 &               91.84 & {\bf 92.07} \\

    CHEMPROT & 82.56 &               82.74 & {\bf 83.49} &                83.4 \\

     QQP-10k & 76.74 &               76.19 &               76.12 & {\bf 79.03} \\

    SCICITE & 81.87 &               82.28 &               81.95 & {\bf 82.74} \\

    MNLI-10k & 65.56 & {\bf 65.74} &               65.31 &               65.14 \\
\midrule
        Mean & 80.26 &               82.47 &               82.67 &               82.74 \\

\bottomrule
\end{tabular}
\caption{Effect of embedding layer to be regularized in \concort.}
\label{tab:concort_layer_by_layer_all}
\end{table}

\clearpage
\section{Hyperparameter Optimization over learning rate, epochs and dropout}\label{sec:hpo}
Table~\ref{tab:hpo_lr_epochs} shows the results when we search over optimal learning rate and number of epochs for each task and method. For learning rate, we perform the search over [5e-6,1e-5,2e-5,4e-5] and for epochs, we search over $[5,10]$. We add an additional baseline method $DR$ where we search over dropout rate from $[0.05,0.1,0.2,0.4]$.
\begin{table}[ht]
\footnotesize
\setlength{\tabcolsep}{2pt}
\centering
\begin{tabular}{lcccccccc}
\toprule
       Tasks & STD++ &   DR&               DA &                  WC & ReInit &   R3F &             \cnrt &             \cprt \\
\midrule
     QQP &  78.6 & 78.9 & 78.8 &  78.2 &  78.7 &  78.6 &  {\bf 79.7} &  79.2 \\
RTE &  72.9 & 73.4 & 74.4 &  74.3 &  74.1 &  74.2 &  {\bf 75.1} &  75.0 \\
        COLA &  56.7 & 61.2& 62.4 &  61.8 & 61.5 &  61.7 &  {\bf 62.5} &  61.8 \\
        MRPC &  90.3 & 90.8 & 90.5 &  90.4 &  90.8 &   90.4 &  90.8 &  {\bf 90.8} \\
\bottomrule
\end{tabular}
\caption{Results with HPO over epochs and learning rate. DR is a baseline method where we do additional HPO over dropout rate as well.}
\label{tab:hpo_lr_epochs}
\end{table}

\section{Experiments for RoBERTa}\label{sec:roberta}
In the results above, we observed that our methods improve significant gain over baseline methods for BERT-large. Table~\ref{tab:robert} shows the result when we compare \concort against \stdpp. We fine-tune the model for 10 epochs with regularization coefficient of 0.01 and learning rate 1e-5. Mean and standard deviation across three runs is reported. We observe that \concort improved \stdpp performance in all cases.

\begin{table}[ht]
\footnotesize
\setlength{\tabcolsep}{2pt}
\centering
\begin{tabular}{lcc}
\toprule
       Tasks & STD++ & \cnrt\\
\midrule
     MRPC & $90.3 \pm 1.0$ & $92.0\pm 0.5$\\
     RTE & $74.1\pm 2.1$ & $77.1\pm 0.7$\\
     CoLA & $60.0\pm 1.1$ & $60.1\pm 0.6$\\
\bottomrule
\end{tabular}
\caption{Results for RoBERTa-base on 3 GLUE datasets.}
\label{tab:robert}
\end{table}

\clearpage
\section{Supervised Contrastive Learning}\label{sec:scl}

Let a mini-batch has $m$ examples, $(x_1,y_1),\dots,(x_m,y_m)$ and $z_1,\dots,z_m$ be the representations (output of encoder) using the model being finetuned. Supervised Contrastive Learning encourages the representations of examples of same label in the mini-batch to be close to each other and far from the examples with different label by additing the following loss to the objective:
\begin{align*}
    \mathcal{L}_{SCL} &= \sum_{i=1}^m  -\frac{1}{N_{y_i}-1} \sum_{j=1}^m 1_{i\neq j} 1_{y_i=y_j} \\
    &\hspace{14mm} \log \frac{exp(\langle z_i,z_j\rangle/\tau)}{\sum_{k=1}^m 1_{i\neq k} exp(\langle z_i, z_k\rangle /\tau)}
\end{align*}
where $\tau$ is a scalable temperature parameter that controls the separation of classes. Loss function during training is 
\[\mathcal{L} = \lambda \mathcal{L}_{CE} + (1-\lambda) \mathcal{L}_{SCL}\]
where $\mathcal{L}_{CE}$ is the cross entropy loss where $\lambda$ is a factor that can be tuned. This was shown to improve finetuning process in~\cite{GunelDCS20} for few-shot finetuning. 

From the definition of $\mathcal{L}_{SCL}$ we observe that SCL is only effective when the mini-batch size is large and each label class is sufficiently represented in the mini-batch. Otherwise, the loss function $\mathcal{L}_{SCL}$ is vacuous. For instance, if the mini-batch size is $1$ which is the case for many of our datasets,  then $\hat{L}_{SCL}=0$ for all the mini-batches. Thus, it is equivalent to the standard finetuning.  Large mini-batch size however requires large memory during finetuning process which is not always available as in our case.Thus, we look for a relaxation of SCL which can be implemented in a memory efficient manner. 

We start by considering $\mathcal{L}_{SCL}$ over the entire input set instead of mini-batch and then replace the example $x_j$ with mean of examples of the same class as $x_j$ while computing similarity with another example. More formally, let the training data be $(x_1,y_1),\dots,(x_N,y_n)$, the set of labels be $\{1,\dots,\ell\}$ and representation of $x_i$ from the encoder of finetuning model. Let $C_{j} = \{i \mid y_i = j\}$ and $c_j = \frac{1}{|C_j|}\sum_{i \in C_j}z_i$ be the center of embeddings of inputs with label $j$. We consider the following relaxation of $\mathcal{L}_{SCL}$. 
\begin{align*} 
\hat{\mathcal{L}}_{SCL}&= \sum_{i=1}^N  -\frac{1}{N_{y_i}-1} \sum_{j=1}^N 1_{i\neq j} 1_{y_i=y_j} \\
    &\hspace{14mm} \log \frac{exp(\langle z_i,c_{y_j}\rangle/\tau)}{\sum_{k=1}^N 1_{i\neq k} exp(\langle z_i, c_{y_k}\rangle /\tau)}\\
    & =-\sum_{i=1}^N \sum_{j=1}^\ell1_{j\neq y_i}\\
    & \hspace{8mm}\log \frac{exp(\langle z_i, c_j\rangle/\tau}{\sum_{k=1}^\ell 1_{k\neq y_i} |C_k| exp(\langle z_i, c_k\rangle/\tau)} 
\end{align*}

A naive implementation of this loss function would be very expansive as the centers $c_1,\dots,c_\ell$ would change in each iteration. We observe that centers change much slower than the individual examples. This is the reason to replace individual training samples with the centers while computing similarity $\langle z_i, z_j\rangle$. Thus, we do not update it in each iteration and instead update it only ten times during the finetuning process. Note that it increases the training time by roughly a factor of $10$ which is also prohibitive for large datasets. Table~\ref{tab:scl_results_250},~\ref{tab:scl_results_500} and ~\ref{tab:scl_results_1000} shows the comparison of memory efficient SCL with our methods. We see that both \concort and \capcort beat SCL consistently for 250, 500 and 1000 training datapoints. Moreover, SCL incur significant loss for several datasets. 
\begin{table}
\footnotesize
\setlength{\tabcolsep}{2pt}
\centering
\begin{tabular}{ccccc}
\toprule
   Tasks &    STD++ &      SCL &  \concort &  \capcort \\

\midrule
\multicolumn{5}{c}{250 datapoints}\\
\midrule
        QNLI & 75.11 &               73.79 & {\bf 78.82} &               76.13 \\

        SST & 88.41 &               87.27 & {\bf 89.29} &               88.59 \\

         QQP & 68.28 &               68.79 &               68.76 & {\bf 70.38} \\

     SCITAIL & 82.31 &               86.13 &  {\bf 88.9} &                86.5 \\

        MNLI &  37.7 &               38.07 &               38.35 & {\bf 41.18} \\

        IMDB & 86.11 &               90.27 & {\bf 90.38} &               90.33 \\

         RTE & 59.13 &               60.77 & {\bf 60.83} &                59.3 \\

        MRPC & 84.43 & {\bf 85.67} &               84.65 &               84.08 \\

        COLA & 41.57 &               31.91 &               43.98 & {\bf 45.28} \\

    CHEMPROT & 55.22 &                32.0 & {\bf 63.28} &               62.32 \\
\midrule
        Mean & 67.83 &               65.47 & {\bf 70.72} &               70.41 \\

Average Rank &   4.3 &                 3.7 &   {\bf 1.6} &                2.35 \\

\bottomrule
\end{tabular}
\caption{Performance of memory-efficiet SCL.}\label{tab:scl_results_250}
\end{table}
\begin{table}
\footnotesize
\setlength{\tabcolsep}{2pt}
\centering
\begin{tabular}{ccccc}
\toprule
   Tasks &    STD++ &      SCL &  \concort &  \capcort \\
\midrule
\multicolumn{5}{c}{500 datapoints}\\
\midrule
        QNLI & 80.86 &               80.48 & {\bf 82.56} &               82.43 \\

        SST & 89.54 & {\bf 90.25} &               89.68 &                89.4 \\

         QQP & 61.12 &               71.58 &               70.97 & {\bf 71.67} \\

     SCITAIL & 85.11 &               87.38 & {\bf 90.07} &                89.3 \\

        MNLI & 41.74 &               44.81 & {\bf 49.07} &               44.73 \\

        IMDB & 83.65 &               90.27 & {\bf 91.32} &               90.49 \\

         RTE & 61.01 & {\bf 64.44} &               62.45 &               62.94 \\

        MRPC & 84.44 &               85.34 &               85.26 & {\bf 85.62} \\

        COLA & 45.29 & {\bf 47.44} &               46.95 &               46.76 \\

    CHEMPROT & 65.59 &               66.55 & {\bf 73.47} &                72.9 \\
\midrule
        Mean & 69.83 &               72.85 & {\bf 74.18} &               73.62 \\

Average Rank &   4.5 &                 2.6 &   {\bf 1.8} &                 2.5 \\
\bottomrule
\end{tabular}
\caption{Performance of memory-efficient SCL.}\label{tab:scl_results_500}
\end{table}
\begin{table}
\footnotesize
\setlength{\tabcolsep}{2pt}
\centering
\begin{tabular}{ccccc}
\toprule
   Tasks &    STD++ &      SCL &  \concort &  \capcort \\
\midrule
\multicolumn{5}{c}{1000 datapoints}\\
\midrule
        QNLI & 81.73 &                83.3 & {\bf 83.67} &               83.25 \\

        SST & 90.69 & {\bf 90.77} &               90.32 &               90.62 \\

         QQP & 70.58 &               72.79 & {\bf 73.04} &               70.67 \\

     SCITAIL & 80.31 &               88.91 & {\bf 91.51} &                90.9 \\

        MNLI & 50.32 &               55.12 &                55.5 & {\bf 56.36} \\

        IMDB &  90.9 &               90.85 & {\bf 91.35} &               91.19 \\

         RTE & 63.63 &               64.98 &               67.06 &                66.7 \\

        MRPC & 87.82 & {\bf 88.01} &               87.54 &               87.97 \\

        COLA & 49.39 &               46.47 & {\bf 51.05} &               49.39 \\

    CHEMPROT & 75.64 &               77.21 & {\bf 79.64} &               79.03 \\
\midrule
        Mean &  74.1 &               75.84 & {\bf 77.07} &               76.61 \\

Average Rank &   4.3 &                 3.1 &   {\bf 1.9} &                2.75 \\
\bottomrule
\end{tabular}
\caption{Performance of memory-efficiet SCL.}\label{tab:scl_results_1000}
\end{table}

\section{Comparison of \concort and \capcort against each baseline}\label{sec:num_wins_comparison}
Table~\ref{tab:num_wins_comparison} show that both \concort and \capcort outperform each baseline method in majority of the datasets/ 
\begin{table}[ht]
\footnotesize
\centering
\setlength{\tabcolsep}{1pt}
\begin{tabular}{ccc}
    \toprule
    
         & \concort & \capcort \\
         \midrule
         \multicolumn{3}{c}{\# wins against baselines methods}\\
    \midrule
         \multicolumn{3}{c}{GLUE datasets (out of 7)}\\
    \midrule
         \stdpp &7 & 6\\
         DA &5 &5\\
         WC & 5 &6\\
         ReInit & 7&7\\
         R3F &7 & 7\\
         \midrule
         \multicolumn{3}{c}{Non-GLUE datasets (out of 6)}\\
    \midrule
         \stdpp &6 & 6\\
         DA &6 &6\\
         WC & 3 &5\\
         ReInit & 6&6\\
         R3F &6 & 6\\
    \bottomrule
    \end{tabular}
\caption{ Number of tasks for which \concort or \capcort outperform the baseline method.
}
\label{tab:num_wins_comparison}
\end{table}

\clearpage
\section{250, 500 and 1000 Training Datapoints}\label{sec:small_datasets_full_results}

Table~\ref{tab:results_250_datapoints},~\ref{tab:results_500_datapoints} and ~\ref{tab:results_1000_datapoints} show the performance for few-sample finetuning setting.
\begin{table*}[ht]
\small
\centering
\begin{tabular}{cccccccc}
\toprule
       Tasks & STD++ &    DA &                  WC &              ReInit &   R3F &             \concort &             \capcort \\

\midrule
\multicolumn{8}{c}{250 Training datapoints}\\
\midrule

         QQP & 68.28 &               65.63 &               69.01 &               68.46 & 67.45 &               68.76 & {\bf 70.38} \\

        COLA & 41.57 &               40.39 &               45.14 &               43.73 & 40.76 &               43.98 & {\bf 45.28} \\

         RTE & 59.13 & {\bf 61.37} &               58.99 &               60.29 & 60.02 &               60.83 &                59.3 \\

        MNLI &  37.7 &               40.24 &               33.71 & {\bf 41.47} &  37.8 &               38.35 &               41.18 \\

        YELP & 92.51 &               92.68 &               92.97 & {\bf 93.64} & 93.13 &               93.13 &                93.3 \\

    CHEMPROT & 55.22 &                58.6 &               59.52 & {\bf 64.92} &  59.6 &               63.28 &               62.32 \\

        QNLI & 75.11 &               78.72 &               75.27 &               77.62 & 77.43 & {\bf 78.82} &               76.13 \\

        IMDB & 86.11 &               89.17 &               89.51 &               89.44 & 89.09 & {\bf 90.38} &               90.33 \\

    SCICITE & 76.86 &                77.2 &               76.23 & {\bf 78.86} & 77.39 &               76.11 &               75.76 \\

        SST & 88.41 &                88.1 &               88.14 &                88.3 & 88.47 & {\bf 89.29} &               88.59 \\

        MRPC & 84.43 &               83.81 & {\bf 84.93} &               84.87 & 84.71 &               84.65 &               84.08 \\

     SCITAIL & 82.31 &               87.65 &                86.2 &                87.8 &  88.6 &  {\bf 88.9} &                86.5 \\

      AGNEWS & 88.08 &               87.51 &               87.55 &               87.36 & 87.99 & {\bf 88.53} &               88.13 \\
\midrule
        Mean & 71.98 &               73.16 &               72.86 & {\bf 74.37} & 73.27 &               74.23 &               73.94 \\

Average Rank &  5.62 &                4.92 &                4.62 &                 3.0 &  4.04 &   {\bf 2.5} &                3.31 \\

\bottomrule
\end{tabular}
\caption{Performance for different regularization methods.}\label{tab:results_250_datapoints}
\end{table*}
\begin{table*}
\small
\centering
\begin{tabular}{cccccccc}
\toprule
       Tasks & STD++ &    DA &                  WC &              ReInit &   R3F &             \concort &             \capcort \\

\midrule

\multicolumn{8}{c}{500 Training datapoints}\\
\midrule
         QQP & 61.12 &                71.5 &               70.13 & {\bf 72.03} &               71.61 &               70.97 &               71.67 \\

        COLA & 45.29 &               46.27 & {\bf 47.34} &               46.03 &               45.36 &               46.95 &               46.76 \\

         RTE & 61.01 &               60.53 &               61.23 & {\bf 63.33} &               60.41 &               62.45 &               62.94 \\

        MNLI & 41.74 &               49.05 &               46.32 & {\bf 51.11} &               42.25 &               49.07 &               44.73 \\

        YELP & 93.32 & {\bf 93.46} &               92.87 &               93.04 &               93.27 &                93.2 &               93.35 \\

    CHEMPROT & 65.59 &  {\bf 74.2} &               70.78 &                73.2 &               70.04 &               73.47 &                72.9 \\

        QNLI & 80.86 &                82.4 &               82.19 &               81.84 &               82.42 & {\bf 82.56} &               82.43 \\

        IMDB & 83.65 &               90.42 & {\bf 91.47} &               90.57 &               89.45 &               91.32 &               90.49 \\

    SCICITE & 78.84 &               79.19 &               79.21 &               79.14 &               78.38 &               79.45 &  {\bf 79.8} \\

        SST & 89.54 &               89.36 &               90.02 & {\bf 90.37} &                90.1 &               89.68 &                89.4 \\

        MRPC & 84.44 &               84.69 &                85.4 &               84.93 &               84.77 &               85.26 & {\bf 85.62} \\

     SCITAIL & 85.11 &               87.41 & {\bf 90.23} &               89.38 &               80.73 &               90.07 &                89.3 \\

      AGNEWS & 89.09 &               88.93 &                89.0 &               89.09 & {\bf 89.59} &               89.53 &               89.54 \\
\midrule
        Mean & 73.81 &               76.72 &               76.63 & {\bf 77.24} &               75.26 &               77.23 &               76.84 \\

Average Rank &  6.08 &                4.38 &                3.69 &                3.31 &                4.85 &  {\bf 2.77} &                2.92 \\

\bottomrule
\end{tabular}
\caption{Performance for different regularization methods.}\label{tab:results_500_datapoints}
\end{table*}
\begin{table*}
\centering
\begin{tabular}{cccccccc}
\toprule
       Tasks & STD++ &    DA &                  WC &              ReInit &   R3F &             \concort &             \capcort \\

\midrule

\multicolumn{8}{c}{1000 Training datapoints}\\
\midrule

         QQP & 70.58 &  71.6 &               70.64 &               71.87 & 71.69 & {\bf 73.04} & 70.67 \\

        COLA & 49.39 & 50.46 &               50.79 &               49.47 & 50.01 & {\bf 51.05} & 49.39 \\

         RTE & 63.63 &  65.7 &               62.09 &               65.78 & 63.54 & {\bf 67.06} &  66.7 \\

        MNLI & 50.32 &  56.1 &               55.28 &  {\bf 57.9} & 55.51 &                55.5 & 56.36 \\

        YELP &  94.3 & 93.83 &               94.41 &               93.99 & 94.51 & {\bf 94.78} & 94.19 \\

    CHEMPROT & 75.64 & 78.88 &               78.37 &               78.12 & 77.98 & {\bf 79.64} & 79.03 \\

        QNLI & 81.73 & 83.46 &               83.51 & {\bf 84.14} & 83.11 &               83.67 & 83.25 \\

        IMDB &  90.9 & 91.22 &               90.99 &               91.29 &  90.8 & {\bf 91.35} & 91.19 \\

    SCICITE & 80.85 & 80.93 & {\bf 81.55} &               80.39 & 80.15 &               80.25 & 81.09 \\

        SST & 90.69 &  90.8 & {\bf 90.81} &               90.67 &  90.5 &               90.32 & 90.62 \\

        MRPC & 87.82 & 87.49 & {\bf 88.12} &               87.29 & 87.92 &               87.54 & 87.97 \\

     SCITAIL & 80.31 & 89.72 & {\bf 91.68} &                91.0 & 91.38 &               91.51 &  90.9 \\

      AGNEWS & 89.29 & 89.38 &               89.25 & {\bf 89.91} & 89.62 &               89.74 & 89.32 \\
\midrule
        Mean & 77.34 &  79.2 &               79.04 &               79.37 & 78.98 & {\bf 79.65} & 79.28 \\

Average Rank &  5.69 &   4.0 &                3.62 &                3.54 &  4.62 &  {\bf 2.69} &  3.85 \\

\bottomrule
\end{tabular}
\caption{Performance for different regularization methods.}\label{tab:results_1000_datapoints}
\end{table*}

\section{Results without any filtered runs}\label{sec:non_filtered_results}
\begin{table}[ht]
\footnotesize
\setlength{\tabcolsep}{2.5pt}
\centering
\begin{tabular}{lccccccc}
\toprule
        & STD++ &                  DA &                  WC & ReInit &   R3F &             \cnrt &             \cprt \\
\midrule
\multicolumn{8}{c}{Successful runs (Failed runs filtered)}\\
\midrule
        Mean & $81.16$ &               $83.74$ &               $83.58$ &   $83.3$ & $82.55$ &               $84.01$ & ${\bf 84.36}$ \\
        std & ${2.87}$ &               ${0.56}$ &               ${0.83}$ &   ${0.76}$ & ${2.19}$ &               ${0.61}$ & ${\bf 0.49}$ \\
        Frac & 0.34 & 0.06 & 0.18 & 0.05 & 0.11 & {\bf 0.00} & 0.04\\
        \midrule
        \multicolumn{8}{c}{All runs (Failed runs not filtered)}\\
        \midrule
        Mean & $64.71$ &               $78.58$ &                $81.4$ &  $79.97$ & $72.97$ & ${\bf 83.99}$ &               $83.04$ \\
        std & ${13.4}$ &               ${6.14}$ &                ${2.89}$ &  ${5.89}$ & ${8.62}$ & ${\bf 0.60}$ &               ${1.28}$ \\
\bottomrule
\end{tabular}
\caption{Stability of fine-tuning results.  \cnrt: \concort, \cprt: \capcort. Frac: fraction of fine-tuning runs filtered due to low performance. Mean, std: mean and standard deviation in performance across all datasets. 
}\label{tab:stability_results}
\end{table}

Table~\ref{tab:results_without_filtering_failed_runs_250},~\ref{tab:results_without_filtering_failed_runs_500},~\ref{tab:results_without_filtering_failed_runs_1000}, and ~\ref{tab:results_without_filtering_failed_runs_all} shows performance without filtering out failed runs.
\begin{table*}
\centering
\begin{tabular}{cccccccc}
\toprule
       Tasks & STD++ &    DA &                  WC &              ReInit &   R3F &             \concort &             \capcort \\

\midrule
\multicolumn{8}{c}{250 datapoints}\\
\midrule
        YELP & 92.51 & 92.68 &               92.97 & {\bf 93.64} & 93.13 &               93.13 &               93.57 \\

         RTE &  55.6 & 58.12 &               55.96 & {\bf 60.29} & 58.88 &               59.78 &               60.14 \\

        MNLI & 25.75 & 30.22 &               28.42 & {\bf 41.47} & 31.94 &                39.2 &               38.58 \\

        QNLI & 72.71 &  77.5 &               75.27 &               77.62 & 77.43 & {\bf 78.82} &               76.13 \\

     SCITAIL & 82.31 & 87.65 &                86.2 &                87.8 &  88.6 &  {\bf 88.9} &                86.5 \\

        SST & 88.41 &  88.1 &               88.14 &                88.3 & 88.47 & {\bf 89.29} &               88.59 \\

    CHEMPROT & 55.22 &  58.6 &               59.52 & {\bf 64.92} &  59.6 &               60.72 &               62.32 \\

      AGNEWS & 88.08 & 87.51 &               87.55 &               87.36 & 87.99 & {\bf 88.53} &               88.13 \\

    SCICITE & 76.86 &  77.2 &               76.23 &               78.86 & 77.39 & {\bf 78.95} &               75.76 \\

        IMDB &  83.5 & 80.13 &               89.51 &               89.44 & 89.05 & {\bf 90.38} &               90.33 \\

        COLA & 41.57 & 40.39 &               45.14 &               43.73 & 40.76 &               43.98 & {\bf 45.28} \\

        MRPC & 84.43 & 83.81 & {\bf 84.93} &               84.87 & 84.71 &               84.65 &               84.08 \\

         QQP & 35.94 & 64.08 &               51.92 &               68.46 & 55.81 & {\bf 68.76} &                65.5 \\
\midrule
        Mean & 67.91 & 71.23 &                70.9 & {\bf 74.37} & 71.83 &               74.24 &               73.45 \\

Average Rank &  5.92 &  5.46 &                4.85 &                2.69 &  3.88 &  {\bf 1.96} &                3.23 \\
\bottomrule
\end{tabular}
\caption{Performance for different regularization methods without filtering failed runs.}\label{tab:results_without_filtering_failed_runs_250}
\end{table*}

\begin{table*}
\centering
\begin{tabular}{cccccccc}
\toprule
       Tasks & STD++ &    DA &                  WC &              ReInit &   R3F &             \concort &             \capcort \\

\midrule
\multicolumn{8}{c}{500 Training Datapoints}\\
\midrule
        YELP & 85.57 & {\bf 93.46} &               92.87 &               93.04 &               93.27 &                93.2 &               93.35 \\

         RTE & 52.47 &               53.86 &               59.81 &               60.87 &               55.64 & {\bf 61.52} &                58.3 \\

        MNLI & 23.87 &               32.25 &               39.53 & {\bf 51.11} &               42.19 &               43.81 &               43.19 \\

        QNLI & 74.76 &               75.45 &               80.64 & {\bf 81.84} &               81.76 &               81.38 &               81.25 \\

     SCITAIL & 85.11 &               87.41 & {\bf 90.23} &               89.38 &               80.73 &               90.07 &                89.3 \\

    CHEMPROT & 65.59 &  {\bf 74.2} &               70.78 &                73.2 &               70.04 &               73.47 &                72.9 \\

        SST & 89.54 &               89.36 &               90.02 & {\bf 90.37} &                90.1 &               89.68 &                89.4 \\

      AGNEWS & 89.09 &               88.93 &                89.0 &               89.09 & {\bf 89.59} &               89.53 &               89.54 \\

    SCICITE & 78.84 &               79.19 &               79.21 &               79.14 &               78.38 &               79.45 &  {\bf 79.8} \\

        IMDB & 78.68 &               90.42 & {\bf 91.47} &               90.57 &               89.45 &               91.32 &               90.49 \\

        COLA & 45.29 &               46.27 & {\bf 47.34} &               46.03 &               45.36 &               46.95 &               46.76 \\

        MRPC & 84.44 &               84.69 &                85.4 &               84.93 &               84.77 &               85.26 & {\bf 85.62} \\

         QQP & 25.73 &                56.3 &               68.63 & {\bf 72.03} &               42.07 &               70.97 &               71.48 \\
\midrule
        Mean & 67.61 &               73.21 &               75.76 & {\bf 77.05} &               72.57 &               76.66 &               76.26 \\

Average Rank &  6.54 &                4.85 &                3.46 &                2.92 &                4.62 &  {\bf 2.54} &                3.08 \\

\bottomrule
\end{tabular}
\caption{Performance for different regularization methods without filtering failed runs.}\label{tab:results_without_filtering_failed_runs_500}
\end{table*}
\begin{table*}
\begin{tabular}{cccccccc}
\toprule
       Tasks & STD++ &    DA &                  WC &              ReInit &   R3F &             \concort &             \capcort \\

\midrule
\multicolumn{8}{c}{1000 Training Datapoints}\\
\midrule
        YELP &  94.3 & 93.83 &               94.24 &               93.99 & 94.51 & {\bf 94.78} & 94.19 \\

         RTE & 49.74 & 53.26 &               59.09 &               62.94 & 52.11 & {\bf 65.78} & 57.16 \\

        MNLI & 32.07 & 31.16 &               53.25 & {\bf 57.38} & 45.06 &               55.88 & 54.73 \\

        QNLI & 69.93 &  82.7 &               82.34 & {\bf 84.14} & 77.63 &               83.67 & 83.23 \\

     SCITAIL & 80.31 & 89.72 & {\bf 91.68} &                91.0 & 91.38 &               91.51 &  90.9 \\

        SST & 90.69 &  90.8 & {\bf 90.81} &               90.67 &  90.5 &               90.75 & 90.62 \\

    CHEMPROT & 75.64 & 78.88 &               78.37 &               78.12 & 77.98 & {\bf 79.64} & 79.03 \\

      AGNEWS & 89.29 & 89.38 &               89.25 & {\bf 89.91} & 89.62 &               89.74 & 89.32 \\

    SCICITE & 80.85 & 80.93 & {\bf 81.55} &               80.39 & 80.82 &               80.25 & 81.09 \\

        IMDB &  82.3 & 91.22 &               90.99 &               91.29 &  90.8 & {\bf 91.35} & 91.19 \\

        COLA & 49.39 & 50.46 &               50.79 &               49.47 & 50.01 & {\bf 51.05} & 49.39 \\

        MRPC & 86.84 & 86.27 & {\bf 88.12} &               87.29 & 87.75 &               87.54 & 87.97 \\

         QQP &  26.0 & 56.63 &               70.26 &               70.36 & 38.79 & {\bf 73.04} & 72.16 \\
\midrule
        Mean &  69.8 & 75.02 &               78.52 &                79.0 & 74.38 & {\bf 79.61} & 78.54 \\

Average Rank &   6.0 &  4.54 &                3.23 &                3.54 &  4.77 &  {\bf 2.15} &  3.77 \\

\bottomrule
\end{tabular}
\caption{Performance for different regularization methods without filtering failed runs.}\label{tab:results_without_filtering_failed_runs_1000}
\end{table*}
\begin{table*}
\begin{tabular}{cccccccc}
\toprule
       Tasks & STD++ &    DA &                  WC &              ReInit &   R3F &             \concort &             \capcort \\

\midrule
\multicolumn{8}{c}{All datapoints}\\
\midrule
        MRPC & 86.84 &               90.67 &                88.6 & 90.98 &  89.9 & {\bf 91.49} &               91.12 \\

    IMDB-10k & 66.59 &               93.24 &               93.69 &  92.7 &  93.1 & {\bf 93.96} &               93.87 \\

    YELP-10k & 72.65 &               95.55 &               95.81 & 95.56 & 95.42 &               95.78 & {\bf 95.96} \\

    SCICITE & 81.87 &               82.13 &               82.23 & 82.41 & 81.97 &               82.74 & {\bf 83.15} \\

    QNLI-10k & 64.93 &               81.64 &                86.1 & 86.73 & 78.43 &               86.85 & {\bf 87.36} \\

    CHEMPROT & 72.59 &               82.57 & {\bf 83.91} & 82.46 & 82.73 &               83.49 &               83.67 \\

    MNLI-10k & 14.37 &               45.98 &               55.94 & 46.54 & 21.66 & {\bf 65.48} &               64.81 \\

        COLA & 59.69 & {\bf 63.45} &                61.5 & 61.25 & 62.04 &               62.34 &               62.47 \\

         RTE & 51.35 &               56.14 &               52.87 & 66.86 & 56.68 & {\bf 71.26} &               61.44 \\

  AGNEWS-10k & 91.67 &               91.82 &               91.92 & 91.67 & 91.73 & {\bf 92.07} &               91.93 \\

        SST & 81.95 &               84.13 &               92.32 & 92.28 & 83.88 &               92.71 & {\bf 93.23} \\

     QQP-10k &   5.9 &               47.77 &               76.25 & 55.16 &  27.8 &               79.03 &  {\bf 79.3} \\

 SCITAIL-10k & 76.01 &               93.74 &               94.03 & 93.36 & 86.54 &               93.74 & {\bf 94.75} \\
\midrule
        Mean & 63.57 &                77.6 &               81.17 & 79.84 & 73.22 & {\bf 83.92} &               83.31 \\

Average Rank &  6.92 &                4.38 &                3.46 &  4.38 &  5.31 &                1.92 &  {\bf 1.62} \\

\bottomrule
\end{tabular}
\caption{Performance for different regularization methods without filtering failed runs.}\label{tab:results_without_filtering_failed_runs_all}
\end{table*}

\section{Detailed results for label noise}\label{sec:detailed_robustness_noise}
Table~\ref{tab:noise_5},~\ref{tab:noise_10},~\ref{tab:noise_20},~\ref{tab:noise_30} shows detailed results with varying amount of label noise in the training data.
\begin{table*}[ht]
\centering
\begin{tabular}{cccccccc}
\toprule
       Tasks & STD++ &    DA &                  WC & ReInit &   R3F &             \concort &             \capcort \\
\midrule
\multicolumn{8}{c}{Noise level = 0.05}\\
\midrule
    SCICITE & 81.29 & 56.46 &               81.34 &               81.11 & 81.03 &               81.36 & {\bf 81.88} \\
    QNLI-10k & 57.35 & 50.18 &               68.59 &               67.16 & 64.33 &               83.87 & {\bf 86.08} \\
        MRPC &  87.4 & 87.48 &               86.66 & {\bf 89.11} & 88.41 &               88.41 &               86.81 \\
         RTE & 51.35 & 48.59 &               51.55 &               65.63 & 48.65 & {\bf 67.58} &               61.16 \\
    IMDB-10k & 72.43 & 68.28 &                77.4 &               62.14 & 91.63 & {\bf 92.84} &               92.56 \\
    CHEMPROT & 71.57 & 81.42 & {\bf 83.88} &               81.81 & 81.48 &               81.64 &               82.18 \\
  AGNEWS-10k &  91.1 & 91.11 & {\bf 91.55} &               91.07 & 91.26 &               91.21 &               91.47 \\
    YELP-10k & 49.92 & 72.44 &               86.24 &                85.4 & 72.38 &                95.1 & {\bf 95.36} \\
    MNLI-10k &   0.0 &   0.0 &                31.9 &               41.99 & 47.51 & {\bf 63.26} &               24.96 \\
    SST-10k & 91.12 & 83.17 & {\bf 91.83} &               90.05 & 91.31 &               83.21 &               90.73 \\
 SCITAIL-10k & 58.21 & 57.94 &               92.91 &               83.36 & 83.93 &               92.32 & {\bf 93.48} \\
     QQP-10k &   0.0 &   0.0 &               76.24 &               57.38 &   0.0 &               77.78 & {\bf 78.82} \\
        COLA & 46.38 & 47.84 &               59.33 &               43.47 & 45.15 & {\bf 59.53} &               48.29 \\
        \midrule
        Mean & 58.32 &  57.3 &               75.34 &               72.28 & 68.24 & {\bf 81.39} &               77.98 \\
Average Rank &   5.5 &  5.96 &                2.92 &                4.38 &  4.35 &  {\bf 2.42} &                2.46 \\
\bottomrule
\end{tabular}
\caption{Training with at most 10k training datapoints on 13 datasets.}
\label{tab:noise_5}
\end{table*}
\begin{table*}[ht]
\centering
\begin{tabular}{cccccccc}
\toprule
       Tasks & STD++ &    DA &                  WC & ReInit &   R3F &             \concort &             \capcort \\
\midrule
\multicolumn{8}{c}{Noise level = 0.1}\\
\midrule
    SST-10k &  88.0 & 66.93 & {\bf 91.58} &               88.97 & 81.02 &               82.73 &               82.64 \\
    MNLI-10k & 12.44 &  0.13 &               57.55 &               48.92 & 24.31 & {\bf 59.67} &               12.38 \\
 SCITAIL-10k & 65.86 & 63.01 &               74.68 &               80.92 & 69.86 &               90.97 & {\bf 92.82} \\
    IMDB-10k & 33.33 & 33.33 &               44.86 &               67.47 & 76.92 &               90.96 & {\bf 91.88} \\
         RTE & 50.54 & 49.46 &               48.38 &               61.52 & 50.54 & {\bf 64.77} &               60.58 \\
        MRPC & 84.11 & 83.69 &               85.11 & {\bf 88.49} & 84.33 &               86.06 &                87.4 \\
  AGNEWS-10k & 90.19 & 90.24 &               90.54 &               90.23 & 90.32 &               90.51 & {\bf 90.58} \\
    CHEMPROT & 80.75 & 68.56 &               82.25 &               80.56 & 69.38 & {\bf 82.49} &               71.92 \\
     QQP-10k &   0.0 &   0.0 &               75.49 &               14.99 & 14.92 &               76.17 & {\bf 76.76} \\
    YELP-10k & 60.78 & 60.77 &               83.65 &                76.4 & 73.85 &               94.22 & {\bf 94.64} \\
        COLA & 56.75 & 45.85 & {\bf 58.65} &                52.8 & 45.43 &               44.12 &                55.6 \\
    QNLI-10k & 50.32 & 50.54 &               65.61 &               63.15 & 66.69 &               74.24 & {\bf 83.83} \\
    SCICITE & 80.79 & 69.21 &               80.45 &               80.71 & 80.29 &               78.76 & {\bf 81.52} \\
    \midrule
        Mean & 57.99 & 52.44 &               72.22 &               68.86 & 63.68 & {\bf 78.13} &               75.58 \\
Average Rank &  4.96 &  6.46 &                3.23 &                3.46 &  4.73 &                2.77 &  {\bf 2.38} \\
\bottomrule
\end{tabular}
\caption{Training with at most 10k training datapoints on 13 datasets.}
\label{tab:noise_10}
\end{table*}
\begin{table*}[ht]
\centering
\begin{tabular}{cccccccc}
\toprule
       Tasks & STD++ &    DA &                  WC & ReInit &   R3F &             \concort &             \capcort \\
\midrule
\multicolumn{8}{c}{Noise level = 0.2}\\
\midrule
    YELP-10k &               49.91 & 49.64 &  {\bf 94.1} &               56.72 & 50.14 &               92.83 &               93.63 \\
         RTE &               48.65 & 47.29 &               50.54 &               55.38 & 47.29 & {\bf 64.98} &               58.84 \\
     QQP-10k &                0.07 &   0.0 &               19.12 &                 0.0 &   0.0 &               72.29 & {\bf 73.52} \\
    CHEMPROT &               70.05 & 70.33 & {\bf 81.22} &               79.75 & 80.03 &               80.44 &               69.08 \\
    IMDB-10k &               33.33 & 33.33 &               45.03 &               33.39 & 33.33 &               77.62 &  {\bf 89.3} \\
  AGNEWS-10k &               85.77 &  86.9 & {\bf 87.54} &               87.52 &  71.6 &                86.6 &                86.8 \\
    SST-10k & {\bf 87.73} & 58.51 &               79.91 &               86.51 & 79.91 &               87.61 &               73.26 \\
    MNLI-10k &                 0.0 &   0.0 & {\bf 48.74} &               10.46 &   0.0 &                 0.0 &                 0.0 \\
    QNLI-10k &               56.13 & 50.27 &               69.56 &               49.89 & 56.04 &               61.93 & {\bf 81.28} \\
        MRPC &               84.45 & 82.25 & {\bf 85.62} &               83.49 & 82.37 &               72.71 &               84.88 \\
 SCITAIL-10k &               49.62 & 49.62 &               74.85 &               49.62 & 49.62 &                69.5 & {\bf 88.91} \\
        COLA &               17.64 &  9.53 &               20.23 & {\bf 47.09} &  19.7 &               20.16 &               35.51 \\
    SCICITE &               45.74 & 38.42 &               77.13 & {\bf 77.57} & 68.46 &               74.41 &               76.89 \\
    \midrule
        Mean &               48.39 & 44.31 &               64.12 &               55.18 & 49.11 &               66.24 & {\bf 70.15} \\
Average Rank &                4.88 &  5.92 &  {\bf 2.19} &                3.58 &  5.27 &                3.31 &                2.85 \\
\bottomrule
\end{tabular}
\caption{Training with at most 10k training datapoints on 13 datasets.}
\label{tab:noise_20}
\end{table*}
\begin{table*}[ht]
\centering
\begin{tabular}{cccccccc}
\toprule
       Tasks & STD++ &    DA &                  WC & ReInit &   R3F &             \concort &             \capcort \\
\midrule
\multicolumn{8}{c}{Noise level = 0.3}\\
\midrule
     QQP-10k &   0.0 &                 0.0 & {\bf 68.62} &               13.08 & 16.01 &               68.31 &               67.36 \\
    SCICITE & 34.18 &               24.67 &               43.99 &  {\bf 73.7} & 62.73 &                67.3 &  {\bf 73.7} \\
        COLA &  3.42 &                6.67 & {\bf 35.97} &               22.14 &  12.8 &               15.01 &               24.99 \\
    IMDB-10k & 33.33 &               33.33 & {\bf 47.15} &               43.14 & 33.33 &               33.33 &                35.6 \\
    MNLI-10k &   0.0 &                 0.0 &  {\bf 1.29} &                 0.0 &   0.0 &                0.22 &                 0.0 \\
    CHEMPROT & 41.96 &                54.1 &               69.21 & {\bf 75.81} & 45.41 &                67.7 &               53.96 \\
    QNLI-10k & 50.11 &               49.89 &               56.27 &               55.13 & 50.27 &               50.18 & {\bf 77.96} \\
  AGNEWS-10k & 70.31 & {\bf 83.14} &               67.91 &               81.75 & 82.41 &               45.87 &               68.06 \\
         RTE & 49.46 &               49.46 &               51.48 &               53.14 & 49.46 & {\bf 60.83} &               54.51 \\
    YELP-10k & 50.02 &               49.56 &               56.02 &               60.22 & 49.79 &               51.99 & {\bf 91.44} \\
 SCITAIL-10k & 49.62 &               49.62 &               58.45 &               49.62 & 49.62 &               56.83 & {\bf 86.71} \\
        MRPC & 80.92 & {\bf 81.38} &               81.26 &               78.42 & 81.22 &               79.54 &               80.63 \\
    SST-10k & 58.23 &               64.04 &               57.73 & {\bf 75.53} & 63.03 &               57.08 &               57.98 \\
    \midrule
        Mean & 40.12 &               41.99 &               53.49 &               52.44 & 45.85 &               50.32 & {\bf 59.45} \\
Average Rank &   5.5 &                4.88 &  {\bf 2.77} &                3.23 &  4.54 &                4.04 &                3.04 \\
\bottomrule
\end{tabular}
\caption{Training with at most 10k training datapoints on 13 datasets.}
\label{tab:noise_30}
\end{table*}

\section{Representation Collapse - Continual learning perspective}~\label{sec:next_task_finetuning}
Table~\ref{tab:next_task_finetuning_qnli},~\ref{tab:next_task_finetuning_qqp},~\ref{tab:next_task_finetuning_mnli},~\ref{tab:next_task_finetuning_agnews},~\ref{tab:next_task_finetuning_imdb},~\ref{tab:next_task_finetuning_sci_cite},~\ref{tab:next_task_finetuning_rte} shows results for representation collapse when we finetune the model for task $A$ using different methods and then finetune the top layer for task $B$.  
\begin{table*}
\centering
\begin{tabular}{llllllll}
\toprule
       Tasks & STD++ &    DA &                  WC & ReInit &                 R3F & \concort &             \capcort \\
\midrule
        COLA & -0.38 & -0.63 &                3.39 &  -1.17 &                 0.0 &    0.93 &  { \bf 5.08 } \\
        MRPC & 81.62 & 81.22 & { \bf 83.13 } &  81.57 &               81.44 &   82.12 &               82.42 \\
     QQP-10k & 29.68 & 28.97 &               43.13 &  31.71 &               22.18 &   46.74 & { \bf 59.26 } \\
    YELP-10k & 50.86 & 51.55 &               55.13 &  51.46 &               50.99 &   52.63 & { \bf 60.04 } \\
 SCITAIL-10k & 59.55 & 58.44 & { \bf 74.46 } &  61.32 &               49.62 &   67.38 &               71.24 \\
    SCICITE & 24.82 & 24.84 & { \bf 28.74 } &  24.75 &               24.67 &   25.02 &               27.71 \\
  AGNEWS-10k & 20.22 &  20.5 &               36.61 &  16.92 &               20.15 &   31.01 &  { \bf 62.9 } \\
    IMDB-10k & 41.36 & 33.33 &               49.87 &   41.0 &               43.36 &   43.57 & { \bf 55.72 } \\
    CHEMPROT &  33.1 & 32.46 & { \bf 33.23 } &  32.92 &               32.34 &   33.09 &               33.15 \\
    MNLI-10k &  8.07 &  8.27 &               16.76 &   8.36 &                6.92 &   14.58 & { \bf 17.69 } \\
    SST-10k & 51.25 & 54.55 &               58.21 &  52.24 &                52.2 &   55.93 & { \bf 66.36 } \\
         RTE & 51.32 & 51.62 &                51.5 &  51.26 & { \bf 53.55 } &   47.83 &               52.17 \\
        \midrule
Mean & 37.62 & 37.09 &               44.51 &  37.69 &               36.45 &   41.74 & { \bf 49.48 } \\
Average Rank &  5.17 &  5.17 &                1.92 &   5.33 &                5.67 &    3.33 &  { \bf 1.42 } \\
\bottomrule
\end{tabular}
\caption{Results for  training top layer for different task after finetuning entire model for  QNLI-10k}
\label{tab:next_task_finetuning_qnli}
\end{table*}
\begin{table*}
\centering
\begin{tabular}{llllllll}
\toprule
       Tasks & STD++ &    DA &                  WC & ReInit &   R3F &             \concort & \capcort \\
\midrule
        MRPC & 81.45 &  81.4 & { \bf 82.67 } &  81.18 & 81.22 &               82.13 &   81.15 \\
    CHEMPROT & 33.46 & 33.46 &               33.46 &  33.46 & 33.46 & { \bf 33.59 } &   33.46 \\
    QNLI-10k & 59.06 & 63.74 & { \bf 67.98 } &  61.65 & 50.54 &               64.95 &   64.58 \\
    YELP-10k & 50.18 & 50.02 &               55.91 &  51.43 & 50.06 & { \bf 64.07 } &   51.32 \\
 SCITAIL-10k & 62.54 & 72.01 &                68.1 &  65.55 & 49.62 & { \bf 76.55 } &   71.55 \\
        COLA &  0.23 & -0.79 &                 0.0 &  -0.01 &   0.0 &  { \bf 5.17 } &   -0.82 \\
  AGNEWS-10k & 16.02 & 19.77 &               32.56 &   19.6 & 10.76 & { \bf 64.32 } &    38.5 \\
    IMDB-10k & 39.92 & 45.49 &               45.45 &  39.17 & 33.33 & { \bf 66.28 } &   34.55 \\
    SCICITE & 24.67 & 24.67 &               24.65 &  24.67 & 24.67 & { \bf 28.51 } &   24.67 \\
    MNLI-10k & 11.73 & 17.18 &               18.91 &  15.72 &   0.0 & { \bf 19.57 } &   19.38 \\
    SST-10k & 49.69 &   nan &               52.24 &  50.89 & 51.03 & { \bf 69.87 } &   58.08 \\
         RTE & 48.86 & 52.71 &               52.89 &  51.44 &  50.9 & { \bf 53.52 } &   52.17 \\
        \midrule
Mean & 39.82 & 41.79 &               44.57 &  41.23 &  36.3 & { \bf 52.38 } &   44.05 \\
Average Rank &  4.96 &  3.79 &                3.08 &   4.79 &  5.67 &  { \bf 1.17 } &    4.04 \\
\bottomrule
\end{tabular}
\caption{Results for  training top layer for different task after finetuning entire model for  QQP-10k}
\label{tab:next_task_finetuning_qqp}
\end{table*}

\begin{table*}
\centering
\begin{tabular}{llllllll}
\toprule
       Tasks &               STD++ &                  DA &    WC &              ReInit &   R3F &             \concort &             \capcort \\
\midrule
         RTE &               51.35 &               56.82 & 52.08 &               67.33 & 56.82 & { \bf 68.95 } &               64.44 \\
    CHEMPROT &               33.46 & { \bf 33.49 } & 33.31 &               33.18 &  33.2 &               33.33 &               33.21 \\
     QQP-10k &                 0.0 &               14.81 & 15.45 &               59.92 & 24.49 & { \bf 60.55 } &               48.03 \\
    QNLI-10k &               50.54 &               51.22 & 52.57 &               53.88 & 52.06 &                52.6 & { \bf 54.51 } \\
 SCITAIL-10k &               49.62 &               56.42 & 57.32 &               78.47 & 68.79 & { \bf 78.58 } &               71.15 \\
        MRPC & { \bf 81.22 } &               80.14 & 81.16 &               81.08 &  80.3 &               80.12 &               79.73 \\
  AGNEWS-10k &                10.0 &               18.45 & 19.69 & { \bf 43.77 } & 21.62 &               34.52 &               28.13 \\
    IMDB-10k &               33.33 &               42.36 & 41.43 & { \bf 57.45 } & 43.13 &               54.98 &               51.87 \\
    SCICITE &               24.67 &               26.03 & 26.02 &               25.69 & 24.65 &               26.63 & { \bf 29.63 } \\
    SST-10k &               50.92 &               55.09 & 61.81 & { \bf 66.49 } & 55.64 &               62.27 &               59.63 \\
    YELP-10k &               50.02 &               52.67 & 54.75 & { \bf 62.68 } &  54.7 &               58.66 &               61.74 \\
        COLA &                 0.0 &                 0.0 &   0.0 &                 0.0 &   0.0 &  { \bf 0.51 } &               -0.69 \\
        \midrule
Mean &               36.26 &               40.63 &  41.3 & { \bf 52.49 } & 42.95 &               50.98 &               48.45 \\
Average Rank &                5.75 &                4.88 &  4.33 &                2.58 &  4.71 &  { \bf 2.25 } &                 3.5 \\
\bottomrule
\end{tabular}
\caption{Results for  training top layer for different task after finetuning entire model for  MNLI-10k}
\label{tab:next_task_finetuning_mnli}
\end{table*}

\begin{table*}
\centering
\begin{tabular}{llllllll}
\toprule
       Tasks & STD++ &    DA &    WC &              ReInit &   R3F &             \concort &             \capcort \\
\midrule
    SCICITE & 33.28 & 33.85 & 36.38 &               29.25 & 31.38 & { \bf 37.88 } &               33.58 \\
        COLA &   1.1 &  5.61 &  4.66 &                3.01 &  3.59 &                8.11 & { \bf 15.16 } \\
    QNLI-10k & 53.98 & 56.17 & 57.89 &               59.45 & 54.23 &                58.6 &  { \bf 61.1 } \\
    YELP-10k & 57.78 &  58.5 & 59.56 &               58.38 & 55.79 & { \bf 61.11 } &               60.11 \\
 SCITAIL-10k &  50.3 & 56.81 & 51.76 &               57.21 & 53.59 &               55.23 & { \bf 67.82 } \\
         RTE & 52.67 & 52.64 &  52.6 & { \bf 54.51 } & 54.24 &               53.13 &               54.39 \\
    IMDB-10k & 57.55 & 60.21 & 59.33 &               57.61 & 56.16 &  { \bf 60.9 } &               59.42 \\
    CHEMPROT & 33.69 & 33.65 & 33.44 &               33.17 & 33.47 & { \bf 34.32 } &               33.54 \\
    SST-10k & 64.64 & 64.43 & 64.99 &                63.9 & 64.46 &               66.34 & { \bf 66.79 } \\
        MRPC & 81.01 & 81.17 & 81.21 & { \bf 81.72 } & 81.11 &                81.3 &               81.04 \\
    MNLI-10k &  5.46 &  5.79 &  6.14 & { \bf 10.54 } &  6.05 &                 7.2 &                9.64 \\
     QQP-10k &  1.95 &  1.19 &  0.05 &               11.24 &   3.2 &                0.51 & { \bf 31.22 } \\
        \midrule
Mean & 41.12 &  42.5 & 42.34 &               43.33 & 41.44 &               43.72 & { \bf 47.82 } \\
Average Rank &  5.58 &  4.17 &  4.42 &                3.83 &  5.17 &                 2.5 &  { \bf 2.33 } \\
\bottomrule
\end{tabular}
\caption{Results for  training top layer for different task after finetuning entire model for  AGNEWS-10k}
\label{tab:next_task_finetuning_agnews}
\end{table*}

\begin{table*}
\centering
\begin{tabular}{llllllll}
\toprule
       Tasks &               STD++ &                  DA &                  WC & ReInit &                 R3F &             \concort &             \capcort \\
\midrule
    CHEMPROT &                 nan &               31.23 &               32.66 &  32.14 &               32.55 & { \bf 33.13 } &               32.94 \\
        COLA &                1.28 &               -0.71 &                4.41 &    0.0 &                 0.0 &                1.93 &  { \bf 7.01 } \\
     QQP-10k &                10.1 &               18.43 &                1.98 &  16.43 & { \bf 22.99 } &                5.02 &               21.81 \\
    QNLI-10k &                52.3 &               54.59 & { \bf 59.36 } &  52.59 &               49.81 &               55.78 &               54.57 \\
    YELP-10k &               91.28 &               91.96 &               78.81 &  83.41 &               91.41 &               93.03 & { \bf 93.44 } \\
 SCITAIL-10k &               51.76 &               69.71 &               61.04 &  49.56 &               51.46 & { \bf 73.47 } &               50.77 \\
         RTE &               53.61 & { \bf 59.21 } &               50.54 &   54.3 &               53.79 &                55.6 &                57.4 \\
  AGNEWS-10k &               31.43 &               37.43 &               40.37 &  23.64 &               29.79 &               44.51 & { \bf 49.81 } \\
    SCICITE & { \bf 32.46 } &               25.24 &               25.13 &  24.67 &               24.67 &               29.07 &               31.67 \\
    MNLI-10k &                6.78 &                9.17 &                7.88 &    7.1 &                7.89 & { \bf 12.17 } &               12.16 \\
    SST-10k &               85.44 &               87.16 &               76.41 &   80.0 &               87.96 & { \bf 89.39 } &               89.11 \\
        MRPC & { \bf 81.22 } &                 nan &               80.56 &  80.64 &               81.17 &               81.17 &               80.65 \\
        \midrule
Mean &               45.24 &               43.95 &               43.26 &  42.04 &               44.46 &               47.86 & { \bf 48.44 } \\
Average Rank &                4.17 &                3.42 &                4.67 &   5.58 &                 4.5 &  { \bf 2.25 } &                2.42 \\
\bottomrule
\end{tabular}
\caption{Results for  training top layer for different task after finetuning entire model for  IMDB-10k}
\label{tab:next_task_finetuning_imdb}
\end{table*}

\begin{table*}
\centering
\begin{tabular}{llllllll}
\toprule
       Tasks &               STD++ &                 DA &                  WC & ReInit &                 R3F &             \concort &             \capcort \\
\midrule
        COLA &                4.89 &               4.86 &                 9.5 &   1.09 &               -0.44 & { \bf 12.78 } &                12.6 \\
    CHEMPROT &               32.29 &              32.09 &               31.95 &  32.17 & { \bf 32.98 } &                31.7 &               31.41 \\
     QQP-10k &                0.49 & { \bf 9.97 } &                0.03 &   6.07 &                2.12 &                 0.0 &                0.26 \\
    QNLI-10k &               57.18 &              58.82 &               60.03 &  56.73 &               53.81 & { \bf 62.24 } &                60.5 \\
    YELP-10k &               62.97 &              65.96 &                63.1 &  62.67 &                57.9 &               67.26 & { \bf 76.47 } \\
 SCITAIL-10k &               49.37 &              49.31 &               57.88 &  49.46 &               49.62 & { \bf 66.74 } &                49.9 \\
         RTE & { \bf 54.66 } &              53.07 &               52.49 &  50.76 &               51.08 &               54.01 &               53.88 \\
  AGNEWS-10k &                48.5 &              49.79 &               55.09 &  38.46 &               34.05 &               57.98 & { \bf 60.96 } \\
    IMDB-10k &               62.83 &              61.85 &               65.36 &  53.94 &               48.51 &               65.12 & { \bf 70.33 } \\
        MRPC &               81.12 &              81.08 & { \bf 81.37 } &  81.12 &               81.12 &               81.02 &               81.28 \\
    MNLI-10k &  { \bf 9.36 } &               7.45 &                8.05 &   8.27 &                 4.1 &                8.21 &                7.32 \\
    SST-10k &                63.3 &              62.58 &               70.41 &  55.46 &                54.3 &               72.05 & { \bf 72.78 } \\
        \midrule
Mean &               43.91 &              44.74 &               46.27 &  41.35 &               39.09 & { \bf 48.26 } &               48.14 \\
Average Rank &                3.75 &               4.42 &                3.42 &    5.0 &                 5.5 &                3.08 &  { \bf 2.83 } \\
\bottomrule
\end{tabular}
\caption{Results for  training top layer for different task after finetuning entire model for  SCICITE}
\label{tab:next_task_finetuning_sci_cite}
\end{table*}

\begin{table*}
\centering
\begin{tabular}{llllllll}
\toprule
       Tasks & STD++ &    DA &    WC &              ReInit &   R3F &             \concort &             \capcort \\
\midrule
        COLA &   0.0 &   0.0 &  1.48 &               -0.59 &   0.0 &  { \bf 5.06 } &                1.91 \\
        MRPC & 81.22 & 81.22 & 81.31 &               80.75 & 81.22 &               81.68 & { \bf 81.91 } \\
     QQP-10k &   0.0 &   0.0 & 18.89 &               60.51 &   0.0 & { \bf 60.87 } &               36.29 \\
    QNLI-10k & 50.54 & 50.54 & 57.24 &               58.24 & 50.54 &  { \bf 62.8 } &               55.56 \\
    YELP-10k & 50.02 & 50.02 & 50.15 &               58.96 & 53.13 & { \bf 63.63 } &                57.1 \\
 SCITAIL-10k & 49.62 & 49.62 & 57.21 &               79.02 & 49.62 & { \bf 82.36 } &               67.64 \\
  AGNEWS-10k &  10.0 &  10.0 & 21.94 &               44.49 & 15.14 & { \bf 52.33 } &               21.44 \\
    IMDB-10k & 33.33 & 33.33 &  38.1 &               56.09 & 35.22 & { \bf 57.18 } &               44.89 \\
    SCICITE & 24.67 & 24.67 & 25.14 &               25.33 & 24.67 & { \bf 31.81 } &               25.96 \\
    CHEMPROT & 33.46 & 33.46 & 33.46 &               33.26 & 33.46 & { \bf 33.56 } &               33.46 \\
    MNLI-10k &   0.0 &   0.0 &   6.4 & { \bf 29.24 } &  2.67 &                27.0 &               20.51 \\
    SST-10k & 50.92 & 50.92 & 55.59 &               60.92 & 53.36 & { \bf 63.88 } &               56.86 \\
        \midrule
Mean & 31.98 & 31.98 & 37.24 &               48.85 & 33.25 & { \bf 51.85 } &               41.96 \\
Average Rank &  5.96 &  5.88 &  3.75 &                3.25 &  5.08 &  { \bf 1.17 } &                2.92 \\
\bottomrule
\end{tabular}
\caption{Results for  training top layer for different task after finetuning entire model for  RTE}
\label{tab:next_task_finetuning_rte}
\end{table*}
\section{Measuring representation collapse}\label{sec:secondary_metrics_detailed}
Table~\ref{tab:eigenvalue_average_top_1},~\ref{tab:eigenvalue_average_top_2},~\ref{tab:eigenvalue_average_top_5},~\ref{tab:eigenvalue_average_top_10},~\ref{tab:eigenvalue_average_top_20} show the sum of top-k normalized eigenvalues (divide each eigenvalue by the sum of eigenvalues) for k=1, 2, 5, 10, and 20. From this, we can observe that almost all the normalized eigenvalues after the first twenty are close to zero

\begin{table*}
\centering
\begin{tabular}{cccccccc}
\toprule
       Tasks &              STD++ &   DA &   WC &             ReInit &                R3F & \concort &            \capcort \\
\midrule
         RTE &  { \bf 1.0 } &  1.0 & 0.98 &               0.94 &                1.0 &    0.91 &               0.98 \\
        MRPC &               0.97 & 0.92 &  0.9 &               0.93 &               0.94 &    0.95 & { \bf 0.99 } \\
    QNLI-10k &               0.97 & 0.97 & 0.96 & { \bf 0.98 } &               0.98 &    0.96 &               0.96 \\
 SCITAIL-10k &               0.98 & 0.95 & 0.96 &               0.96 &               0.97 &    0.96 & { \bf 0.98 } \\
    IMDB-10k &               0.97 & 0.97 & 0.98 & { \bf 0.98 } &               0.98 &    0.96 &               0.97 \\
    SST-10k &               0.94 & 0.94 & 0.95 &               0.96 &               0.94 &    0.95 & { \bf 0.97 } \\
        COLA &               0.94 & 0.94 & 0.93 &               0.94 & { \bf 0.96 } &    0.94 &               0.95 \\
  AGNEWS-10k &               0.65 & 0.66 & 0.66 & { \bf 0.68 } &               0.66 &    0.65 &               0.64 \\
     QQP-10k &               0.98 & 0.97 & 0.94 &               0.98 &  { \bf 1.0 } &    0.93 &               0.97 \\
    MNLI-10k & { \bf 0.98 } & 0.93 & 0.94 &               0.89 &               0.97 &    0.85 &               0.91 \\
    YELP-10k &               0.98 & 0.97 & 0.97 &               0.99 & { \bf 0.99 } &    0.97 &               0.98 \\
    CHEMPROT &               0.59 & 0.49 & 0.49 &               0.52 &               0.51 &    0.51 & { \bf 0.68 } \\
    SCICITE &               0.86 & 0.87 & 0.87 & { \bf 0.92 } &                0.9 &    0.84 &               0.86 \\
\bottomrule
\end{tabular}
\caption{Normalized average of top-1 eigenvalues}
\label{tab:eigenvalue_average_top_1}
\end{table*}

\begin{table*}
\centering
\begin{tabular}{cccccccc}
\toprule
       Tasks &              STD++ &   DA &   WC &             ReInit &                R3F & \concort &            \capcort \\
\midrule
         RTE &  { \bf 1.0 } &  1.0 & 0.99 &               0.97 &                1.0 &    0.95 &               0.99 \\
        MRPC &               0.98 & 0.96 & 0.95 &               0.94 &               0.98 &    0.97 & { \bf 0.99 } \\
    QNLI-10k &               0.99 & 0.99 & 0.98 & { \bf 0.99 } &               0.99 &    0.98 &               0.98 \\
 SCITAIL-10k & { \bf 0.99 } & 0.98 & 0.98 &               0.99 &               0.99 &    0.98 &               0.99 \\
    IMDB-10k &               0.99 & 0.99 & 0.99 & { \bf 0.99 } &               0.99 &    0.98 &               0.99 \\
    SST-10k &               0.97 & 0.97 & 0.97 &               0.98 &               0.97 &    0.98 & { \bf 0.98 } \\
        COLA &               0.97 & 0.96 & 0.96 &               0.97 & { \bf 0.98 } &    0.98 &               0.97 \\
  AGNEWS-10k &                0.9 &  0.9 &  0.9 & { \bf 0.95 } &               0.91 &    0.91 &               0.81 \\
     QQP-10k &                1.0 & 0.99 & 0.98 &               0.99 &  { \bf 1.0 } &    0.96 &               0.99 \\
    MNLI-10k & { \bf 0.99 } & 0.96 & 0.98 &               0.96 &               0.98 &    0.93 &               0.97 \\
    YELP-10k &               0.99 & 0.99 & 0.98 &                1.0 &  { \bf 1.0 } &    0.99 &               0.99 \\
    CHEMPROT &                0.7 & 0.61 & 0.61 &               0.64 &               0.64 &    0.63 & { \bf 0.77 } \\
    SCICITE &               0.92 & 0.92 & 0.92 & { \bf 0.96 } &               0.95 &     0.9 &               0.92 \\
\bottomrule
\end{tabular}
\caption{Normalized average of top-2 eigenvalues}
\label{tab:eigenvalue_average_top_2}
\end{table*}

\begin{table*}
\centering
\begin{tabular}{cccccccc}
\toprule
       Tasks &             STD++ &   DA &   WC &             ReInit &                R3F & \concort &            \capcort \\
\midrule
         RTE & { \bf 1.0 } &  1.0 &  1.0 &               0.99 &                1.0 &    0.98 &                1.0 \\
        MRPC &              0.99 & 0.99 & 0.98 &               0.95 &               0.99 &    0.99 &  { \bf 1.0 } \\
    QNLI-10k &               1.0 &  1.0 & 0.99 &  { \bf 1.0 } &                1.0 &     1.0 &               0.99 \\
 SCITAIL-10k &               1.0 & 0.99 & 0.99 &  { \bf 1.0 } &                1.0 &     1.0 &                1.0 \\
    IMDB-10k &               1.0 &  1.0 &  1.0 &  { \bf 1.0 } &                1.0 &    0.99 &                1.0 \\
    SST-10k &              0.99 & 0.99 & 0.99 &  { \bf 1.0 } &               0.99 &    0.99 &                1.0 \\
        COLA &              0.99 & 0.99 & 0.98 &               0.99 &  { \bf 1.0 } &    0.99 &               0.99 \\
  AGNEWS-10k &              0.97 & 0.97 & 0.96 & { \bf 0.99 } &               0.97 &    0.97 &               0.97 \\
     QQP-10k &               1.0 &  1.0 & 0.99 &                1.0 &  { \bf 1.0 } &    0.99 &                1.0 \\
    MNLI-10k & { \bf 1.0 } & 0.99 & 0.99 &               0.99 &                1.0 &    0.98 &               0.99 \\
    YELP-10k &               1.0 &  1.0 & 0.99 &  { \bf 1.0 } &                1.0 &     1.0 &                1.0 \\
    CHEMPROT &              0.89 & 0.85 & 0.85 &               0.88 &               0.87 &    0.86 & { \bf 0.92 } \\
    SCICITE &              0.98 & 0.97 & 0.97 & { \bf 0.99 } &               0.99 &    0.97 &               0.98 \\
\bottomrule
\end{tabular}
\caption{Normalized average of top-5 eigenvalues}
\label{tab:eigenvalue_average_top_5}
\end{table*}

\begin{table*}
\centering
\begin{tabular}{cccccccc}
\toprule
       Tasks &             STD++ &   DA &   WC &             ReInit &                R3F & \concort &            \capcort \\
\midrule
         RTE & { \bf 1.0 } &  1.0 &  1.0 &               0.99 &                1.0 &    0.99 &                1.0 \\
        MRPC &               1.0 & 0.99 & 0.99 &               0.96 &                1.0 &     1.0 &  { \bf 1.0 } \\
    QNLI-10k &               1.0 &  1.0 &  1.0 &  { \bf 1.0 } &                1.0 &     1.0 &                1.0 \\
 SCITAIL-10k & { \bf 1.0 } &  1.0 &  1.0 &                1.0 &                1.0 &     1.0 &                1.0 \\
    IMDB-10k &               1.0 &  1.0 &  1.0 &  { \bf 1.0 } &                1.0 &     1.0 &                1.0 \\
    SST-10k &               1.0 &  1.0 & 0.99 &  { \bf 1.0 } &                1.0 &     1.0 &                1.0 \\
        COLA &              0.99 & 0.99 & 0.98 &               0.99 &  { \bf 1.0 } &     1.0 &                1.0 \\
  AGNEWS-10k &              0.98 & 0.98 & 0.98 & { \bf 0.99 } &               0.98 &    0.98 &               0.99 \\
     QQP-10k &               1.0 &  1.0 &  1.0 &                1.0 &  { \bf 1.0 } &    0.99 &                1.0 \\
    MNLI-10k & { \bf 1.0 } & 0.99 & 0.99 &               0.99 &                1.0 &    0.99 &                1.0 \\
    YELP-10k &               1.0 &  1.0 &  1.0 &  { \bf 1.0 } &                1.0 &     1.0 &                1.0 \\
    CHEMPROT &              0.98 & 0.97 & 0.97 &               0.98 &               0.98 &    0.98 & { \bf 0.99 } \\
    SCICITE &              0.99 & 0.99 & 0.98 &  { \bf 1.0 } &                1.0 &    0.98 &               0.99 \\
\bottomrule
\end{tabular}
\caption{Normalized average of top-10 eigenvalues}
\label{tab:eigenvalue_average_top_10}
\end{table*}

\begin{table*}
\centering
\begin{tabular}{cccccccc}
\toprule
       Tasks &             STD++ &                DA &   WC &             ReInit &                R3F &           \concort &           \capcort \\
\midrule
         RTE & { \bf 1.0 } &               1.0 &  1.0 &                1.0 &                1.0 &              0.99 &               1.0 \\
        MRPC &               1.0 &               1.0 &  1.0 &               0.97 &                1.0 &               1.0 & { \bf 1.0 } \\
    QNLI-10k &               1.0 &               1.0 &  1.0 &  { \bf 1.0 } &                1.0 &               1.0 &               1.0 \\
 SCITAIL-10k & { \bf 1.0 } &               1.0 &  1.0 &                1.0 &                1.0 &               1.0 &               1.0 \\
    IMDB-10k &               1.0 &               1.0 &  1.0 &  { \bf 1.0 } &                1.0 &               1.0 &               1.0 \\
    SST-10k &               1.0 &               1.0 & 0.99 &  { \bf 1.0 } &                1.0 &               1.0 &               1.0 \\
        COLA &               1.0 &               1.0 & 0.99 &               0.99 &  { \bf 1.0 } &               1.0 &               1.0 \\
  AGNEWS-10k &              0.99 &              0.99 & 0.98 & { \bf 0.99 } &               0.99 &              0.99 &              0.99 \\
     QQP-10k &               1.0 &               1.0 &  1.0 &                1.0 &  { \bf 1.0 } &               1.0 &               1.0 \\
    MNLI-10k & { \bf 1.0 } &               1.0 &  1.0 &                1.0 &                1.0 &               1.0 &               1.0 \\
    YELP-10k &               1.0 &               1.0 &  1.0 &  { \bf 1.0 } &                1.0 &               1.0 &               1.0 \\
    CHEMPROT &              0.99 &              0.99 & 0.98 &               0.99 &               0.99 &              0.99 & { \bf 1.0 } \\
    SCICITE &              0.99 &              0.99 & 0.99 &  { \bf 1.0 } &                1.0 &              0.99 &              0.99 \\
\bottomrule
\end{tabular}
\caption{Normalized average of top-20 eigenvalues}
\label{tab:eigenvalue_average_top_20}
\end{table*}

Table~\ref{tab:gm_top_5},~\ref{tab:gm_top_10} and ~\ref{tab:gm_top_20} show the \firdivmeasure-k for k=5, 10 and 20. Table~\ref{tab:hm_top_5},~\ref{tab:hm_top_10},~\ref{tab:hm_top_20} show the \secdivmeasure-k for k=5, 10 and 20. We observe that \concort achieves the highest value and thus is most effective in reducing representation collapse. 
\begin{table*}
\centering
\begin{tabular}{cccccccc}
\toprule
       Tasks &   STD++ &                   DA &                   WC &  ReInit &                   R3F &              \concort &               \capcort \\
\midrule
         RTE &    0.06 &                 0.05 &               151.08 &  356.83 &                  4.78 & { \bf 497.38 } &                140.33 \\
        MRPC &  248.44 &               489.14 & { \bf 573.15 } &  254.92 &                437.79 &               368.16 &                181.92 \\
    QNLI-10k &  198.62 &               192.21 &               296.25 &  144.09 &                197.74 & { \bf 316.57 } &                296.03 \\
 SCITAIL-10k &  182.57 & { \bf 369.51 } &               263.48 &  252.59 &                234.66 &               354.42 &                252.86 \\
    IMDB-10k &  248.12 &               244.33 &               179.67 &  142.64 &                211.15 & { \bf 299.57 } &                293.71 \\
    SST-10k &  365.78 &               400.84 &               333.75 &  281.04 &  { \bf 403.95 } &               360.22 &                348.41 \\
        COLA &  405.79 & { \bf 444.48 } &               406.54 &  301.56 &                292.11 &               413.04 &                402.62 \\
  AGNEWS-10k & 1124.36 &              1126.63 &              1083.96 &  818.83 &               1099.94 &              1121.69 & { \bf 1272.38 } \\
     QQP-10k &  109.81 &               163.71 &               380.98 &  211.71 &                  0.07 & { \bf 497.27 } &                234.59 \\
    MNLI-10k &  119.46 &               393.29 &                287.3 &  671.92 &                175.47 & { \bf 849.31 } &                432.63 \\
    YELP-10k &  177.91 &               244.31 &               238.38 &   97.23 &                 87.91 & { \bf 259.35 } &                198.54 \\
    CHEMPROT & 1103.13 &              1351.65 &              1347.29 & 1369.47 & { \bf 1390.28 } &              1352.48 &               1193.84 \\
    SCICITE &   867.7 &               844.45 &               752.98 &  630.59 &                633.57 & { \bf 906.15 } &                775.52 \\
        \midrule
Mean &  396.29 &               481.89 &               484.22 &  425.65 &                397.65 & { \bf 584.28 } &                463.33 \\
Average Rank &    4.85 &                 3.31 &                 3.85 &    5.15 &                  4.92 &    { \bf 2.0 } &                  3.92 \\
\bottomrule
\end{tabular}
\caption{\firdivmeasure-5}
\label{tab:gm_top_5}
\end{table*}

\begin{table*}
\centering
\begin{tabular}{cccccccc}
\toprule
       Tasks &  STD++ &                   DA &                   WC & ReInit &    R3F &              \concort & \capcort \\
\midrule
         RTE &   0.01 &                 0.01 &                28.58 &  54.68 &   0.54 & { \bf 113.27 } &   24.64 \\
        MRPC &  47.83 &                85.21 & { \bf 108.67 } &  65.41 &   59.4 &                61.13 &   19.25 \\
    QNLI-10k &   29.3 &                34.28 &  { \bf 54.86 } &  16.92 &  29.65 &                47.15 &   46.11 \\
 SCITAIL-10k &  25.23 &  { \bf 58.43 } &                47.44 &  26.62 &  31.48 &                49.61 &   33.93 \\
    IMDB-10k &  32.82 &                32.53 &                29.47 &  12.69 &  23.31 &  { \bf 48.51 } &   35.45 \\
    SST-10k &  58.83 &  { \bf 68.08 } &                65.03 &  27.32 &   67.7 &                59.49 &   44.76 \\
        COLA &  69.32 &                82.06 &  { \bf 84.79 } &  44.59 &  42.05 &                68.46 &   71.23 \\
  AGNEWS-10k & 220.03 & { \bf 221.59 } &               186.54 &   95.6 & 202.22 &                208.1 &   179.2 \\
     QQP-10k &   10.5 &                 25.4 &                64.85 &  22.53 &   0.01 &  { \bf 90.66 } &   35.88 \\
    MNLI-10k &  20.29 &                82.66 &                46.99 &  115.1 &   34.8 & { \bf 164.62 } &   71.13 \\
    YELP-10k &  19.27 &                33.91 &   { \bf 39.9 } &   8.13 &  10.12 &                33.49 &   25.33 \\
    CHEMPROT & 499.58 & { \bf 646.59 } &                613.5 & 603.17 & 601.67 &                619.9 &  446.36 \\
    SCICITE &  164.7 &               171.53 &               163.86 &  88.84 &  108.6 & { \bf 190.12 } &  152.38 \\

        \midrule
Mean &  92.13 &               118.64 &               118.04 &  90.89 &   93.2 & { \bf 134.96 } &    91.2 \\
Average Rank &   5.08 &                 2.62 &                 2.85 &   5.38 &   5.31 &   { \bf 2.31 } &    4.46 \\
\bottomrule
\end{tabular}
\caption{\firdivmeasure-10}
\label{tab:gm_top_10}
\end{table*}

\begin{table*}
\centering
\begin{tabular}{cccccccc}
\toprule
       Tasks &              STD++ &                  DA &                  WC &             ReInit &   R3F &             \concort & \capcort \\
\midrule
         RTE &                0.0 &                 0.0 &                5.75 &              12.17 &  0.09 & { \bf 24.83 } &    4.34 \\
        MRPC &               7.55 &               15.06 &               20.88 & { \bf 24.8 } &  8.25 &               10.41 &    2.65 \\
    QNLI-10k &               4.01 &                5.22 &  { \bf 9.67 } &                2.0 &   4.2 &                6.66 &     7.5 \\
 SCITAIL-10k &               2.89 &                7.58 &   { \bf 8.7 } &               3.04 &  3.58 &                5.72 &    4.64 \\
    IMDB-10k &               3.38 &                3.55 &                5.36 &               1.03 &  2.38 &  { \bf 7.34 } &    4.65 \\
    SST-10k &               9.62 &               11.93 & { \bf 14.43 } &               3.63 & 11.54 &               10.26 &    6.77 \\
        COLA &              12.85 &               14.76 &  { \bf 20.1 } &              10.88 &  6.18 &               11.65 &   12.35 \\
  AGNEWS-10k & { \bf 38.7 } &               38.02 &               37.23 &              15.59 & 35.16 &               36.44 &   32.92 \\
     QQP-10k &               0.89 &                2.69 &               10.25 &               2.28 &   0.0 & { \bf 16.01 } &     5.4 \\
    MNLI-10k &               3.44 &               15.14 &                9.09 &              16.15 &  6.35 & { \bf 29.53 } &    12.0 \\
    YELP-10k &               1.78 &                3.68 &  { \bf 6.86 } &               0.54 &  1.05 &                4.18 &    3.36 \\
    CHEMPROT &              69.87 & { \bf 93.37 } &               81.62 &              65.05 & 80.71 &               86.34 &   48.23 \\
    SCICITE &              29.61 &                30.0 &               32.32 &              12.79 & 17.54 & { \bf 37.96 } &   27.58 \\
        \midrule
Mean &               14.2 &               18.54 &               20.17 &              13.07 & 13.62 &  { \bf 22.1 } &   13.26 \\
Average Rank &               5.08 &                3.08 &  { \bf 2.08 } &               5.38 &  5.38 &                2.46 &    4.54 \\
\bottomrule
\end{tabular}
\caption{\firdivmeasure-20}
\label{tab:gm_top_20}
\end{table*}

\begin{table*}
\centering
\begin{tabular}{cccccccc}
\toprule
       Tasks &  STD++ &                   DA &                   WC & ReInit &                  R3F &              \concort &              \capcort \\
\midrule
         RTE &    0.0 &                  0.0 &                46.98 & 129.47 &                 0.63 &  { \bf 210.7 } &                50.95 \\
        MRPC &  95.55 &               198.64 & { \bf 228.53 } &  75.07 &               138.94 &               127.19 &                29.27 \\
    QNLI-10k &  69.03 &                 67.1 &               106.82 &  36.64 &                58.54 & { \bf 108.56 } &                88.44 \\
 SCITAIL-10k &  52.56 & { \bf 123.99 } &               105.75 &  61.73 &                74.35 &               111.24 &                71.91 \\
    IMDB-10k &  69.28 &                67.86 &                60.84 &  24.22 &                58.72 &  { \bf 90.53 } &                 64.7 \\
    SST-10k & 118.65 &               145.93 &               124.11 &  47.52 & { \bf 153.48 } &               129.43 &                 90.6 \\
        COLA & 122.08 & { \bf 171.42 } &               157.71 &  63.75 &                95.61 &               127.75 &                133.4 \\
  AGNEWS-10k & 540.72 &               570.89 &               495.34 & 179.74 &               514.01 &               555.69 & { \bf 702.69 } \\
     QQP-10k &  21.54 &                55.95 &               125.51 &  54.76 &                  0.0 & { \bf 172.94 } &                60.37 \\
    MNLI-10k &  59.91 &               193.92 &                84.33 & 262.68 &                84.15 & { \bf 416.25 } &               166.81 \\
    YELP-10k &  37.49 &  { \bf 78.53 } &                78.31 &  17.42 &                19.99 &                64.93 &                44.16 \\
    CHEMPROT & 980.72 &              1196.85 &              1201.61 & 1199.8 & { \bf 1222.6 } &              1187.46 &                924.5 \\
    SCICITE & 414.13 &               418.66 &               338.99 & 250.67 &               272.92 &  { \bf 476.9 } &               394.16 \\
        \midrule
Mean & 198.59 &               253.06 &               242.68 & 184.88 &               207.23 & { \bf 290.74 } &               217.07 \\
Average Rank &    5.0 &                 2.85 &                 3.31 &   5.62 &                 4.77 &   { \bf 2.31 } &                 4.15 \\
\bottomrule
\end{tabular}
\caption{\secdivmeasure-5}
\label{tab:hm_top_5}
\end{table*}

\begin{table*}
\centering
\begin{tabular}{cccccccc}
\toprule
       Tasks &  STD++ &                   DA &                  WC & ReInit &    R3F &             \concort & \capcort \\
\midrule
         RTE &    0.0 &                  0.0 &                 9.0 &  14.79 &    0.1 & { \bf 41.73 } &    7.15 \\
        MRPC &  15.09 &                24.73 & { \bf 33.48 } &   26.9 &  13.53 &               17.29 &    3.31 \\
    QNLI-10k &   6.99 &                 9.65 & { \bf 16.49 } &   3.13 &    7.1 &                11.6 &   12.37 \\
 SCITAIL-10k &   5.46 &  { \bf 14.88 } &               14.11 &   4.35 &   6.74 &               10.75 &    7.91 \\
    IMDB-10k &    6.7 &                 6.83 &                7.81 &    1.7 &   4.09 & { \bf 11.92 } &    6.44 \\
    SST-10k &  15.31 &                 19.3 & { \bf 21.21 } &   4.19 &  18.37 &               16.18 &    9.32 \\
        COLA &  19.72 &                24.21 & { \bf 28.76 } &   11.5 &   9.88 &               19.34 &   20.15 \\
  AGNEWS-10k &  65.88 &  { \bf 67.33 } &               54.52 &  18.78 &  60.41 &               62.36 &   44.45 \\
     QQP-10k &   1.56 &                  5.6 &               17.67 &   3.68 &    0.0 & { \bf 26.38 } &    9.33 \\
    MNLI-10k &   5.68 &                27.28 &                13.1 &  28.45 &  10.77 & { \bf 50.47 } &   20.74 \\
    YELP-10k &   3.28 &                 7.42 &  { \bf 11.0 } &   0.95 &   1.88 &                7.14 &    4.95 \\
    CHEMPROT & 302.28 & { \bf 429.53 } &              333.57 & 356.82 & 332.11 &              389.25 &  226.06 \\
    SCICITE &  48.68 &                 54.7 &               56.06 &  18.57 &  29.54 & { \bf 59.65 } &   45.86 \\
        \midrule
Mean &   38.2 &                53.19 &               47.44 &  37.99 &  38.04 &  { \bf 55.7 } &   32.16 \\
Average Rank &   5.08 &                 2.77 &  { \bf 2.31 } &   5.31 &   5.46 &                2.46 &    4.62 \\
\bottomrule
\end{tabular}
\caption{\secdivmeasure-10}
\label{tab:hm_top_10}
\end{table*}

\begin{table*}
\centering
\begin{tabular}{cccccccc}
\toprule
       Tasks & STD++ &                  DA &                  WC &              ReInit &   R3F &             \concort & \capcort \\
\midrule
         RTE &   0.0 &                 0.0 &                1.93 &                4.48 &  0.02 &  { \bf 9.05 } &    1.24 \\
        MRPC &  2.04 &                 4.6 &                6.63 & { \bf 13.83 } &  2.13 &                3.05 &    0.61 \\
    QNLI-10k &   0.9 &                1.31 &  { \bf 2.96 } &                0.39 &  0.98 &                 1.5 &    2.04 \\
 SCITAIL-10k &  0.55 &                1.53 &  { \bf 2.65 } &                 0.6 &  0.68 &                1.04 &     1.1 \\
    IMDB-10k &  0.52 &                0.62 &                1.63 &                0.13 &  0.37 &  { \bf 1.84 } &    1.02 \\
    SST-10k &  2.61 &                3.52 &  { \bf 5.33 } &                0.83 &  3.29 &                2.94 &    1.73 \\
        COLA &  3.99 &                4.49 &  { \bf 7.88 } &                4.37 &  1.52 &                3.28 &     3.6 \\
  AGNEWS-10k & 11.43 &               11.09 & { \bf 12.51 } &                4.21 & 10.29 &               10.72 &   10.49 \\
     QQP-10k &  0.12 &                0.43 &                2.74 &                0.38 &   0.0 &  { \bf 4.67 } &    1.39 \\
    MNLI-10k &  0.94 &                4.56 &                2.93 &                3.71 &  1.87 &  { \bf 8.72 } &     3.4 \\
    YELP-10k &  0.25 &                0.61 &  { \bf 1.96 } &                0.06 &  0.17 &                0.91 &    0.74 \\
    CHEMPROT & 14.88 & { \bf 20.84 } &               17.98 &               12.33 & 17.11 &               17.93 &    6.92 \\
    SCICITE &  8.79 &                8.87 &               10.77 &                3.06 &  4.68 & { \bf 12.55 } &    8.44 \\
        \midrule
Mean &  3.62 &                4.81 &                5.99 &                3.72 &  3.32 &  { \bf 6.02 } &    3.29 \\
Average Rank &  5.23 &                3.15 &  { \bf 1.85 } &                5.23 &  5.46 &                2.69 &    4.38 \\
\bottomrule
\end{tabular}
\caption{\secdivmeasure-20}
\label{tab:hm_top_20}
\end{table*}

\section{Walltime Analysis}\label{sec:walltime}
\paragraph{Walltime analysis:} \stdpp uses a single forward and backward pass with simplest loss function and thus has the least training time. ReInit is a close second as it only differs in the initialization of the model. WC also uses a single forward and backward pass but is slower due to the regularization loss function computation. R3F and DA use two forward passes and two (effective) backward passes. Our method on the other hand use only one forward and backward pass. In addition to that we use only an extra forward pass of the pretrained model.
Thus, our method is slower than \stdpp, ReInit and WC and is faster than R3F and DA. 
Table~\ref{tab:wall_time_analysis} show the training time for all the methods. We observe that R3F consistently  takes more time than all the methods. \concort runs faster than R3F and DA but slower than \stdpp, WC and ReInit. \capcort runs slower than \concort.
\begin{table*}
\centering
\begin{tabular}{llllllll}
\toprule
       Tasks &   STD++ &      DA &      WC &  ReInit &                    R3F & \concort & \capcort \\
\midrule
    CHEMPROT &  584.01 & 1015.37 &  702.38 &  589.39 &           1415.28 &  826.09 & 1141.63 \\
    MNLI-10k & 1506.74 & 2643.21 & 1723.52 & 1514.14 &           3746.99 & 1832.83 & 3022.97 \\
 SCITAIL-10k & 1678.89 & 2962.31 &  1896.3 & 1684.06 &           4217.93 & 2418.12 & 3130.63 \\
        MRPC &  162.09 &  258.92 &  214.94 &  157.42 &            362.83 &  226.59 &  299.13 \\
    QNLI-10k & 1683.58 & 2966.56 & 1894.97 & 1681.34 &           4232.07 & 2414.78 & 3124.71 \\
     QQP-10k & 1272.46 & 2194.95 & 1487.52 & 1276.92 &           3078.56 & 2307.62 & 2302.41 \\
        SST & 5050.32 & 8537.84 & 5997.04 & 5080.05 &          11878.49 &  6019.1 & 9568.79 \\
        COLA &  239.78 &  375.78 &  349.94 &  240.61 &            515.27 &  292.32 &   398.9 \\
    SCICITE &  961.03 & 1746.23 & 1066.78 &   962.4 &           2838.07 & 1850.68 &  1835.6 \\
    IMDB-10k & 1679.74 & 2975.29 & 1899.58 & 1692.33 &            4220.9 &  3149.3 &  3124.5 \\
    YELP-10k & 1681.65 & 2975.09 & 1894.85 & 1686.21 &           4224.52 & 2426.62 & 3130.62 \\
         RTE &  307.13 &   504.8 &  396.35 &  309.26 &            698.16 &  527.76 &  529.66 \\
  AGNEWS-10k &  589.23 & 1046.27 &  675.37 &  595.06 &            1449.9 & 1125.27 & 1227.09 \\
  \midrule
        Mean &  1338.2 & 2323.28 & 1553.81 & 1343.78 &           3298.38 & 1955.16 &  2525.9 \\
\bottomrule
\end{tabular}
\caption{Training time for different methods}
\label{tab:wall_time_analysis}
\end{table*}
\section{Connection of \firdivmeasure and \secdivmeasure to parameter estimation error}\label{sec:our_metric_parameter_estimation}
Let the pseudo linear regression task on finetuned representations be defined by $w \in \rR^d$ and the noisy labels observed on $z_i$'s be $y_i = z_i^T w + \epsilon_i$ where $\epsilon_i$'s are the gaussian noise centered around ${\bf 0}$ with identity covariance matrix. If $\hat{w}$ is the least square minimizer (same as log-likelihood maximizer), then
$\hat{w} = w + N\left(0, G^{-1}\right)$

\firdivmeasure\space corresponds to minimizing the confidence ellipsoid corresponding to the error $\hat{w}-w$. \secdivmeasure\space corresponds to minimizing the expected $\ell_2^2$ norm of the error vector $\hat{w} - w$. Derivation of the $\hat{w}$ and the explanation can be found in ~\citet{MadanSTX19}.

\end{document}